\documentclass{workingpaperJJB}
\usepackage{setspace}   
\usepackage{xcolor}
\usepackage{hyperref}
\RequirePackage{tgtermes}
\RequirePackage{newtxtext}
\RequirePackage{newtxmath}
\RequirePackage{bm}
\RequirePackage{endnotes}
\DeclareMathOperator{\diam}{diam}
\DeclareMathOperator{\var}{Var}
\DeclareMathOperator{\cov}{Cov}

\OneAndAHalfSpacedXII 

\usepackage{algorithm}
\usepackage{algpseudocode}
\usepackage{tikz}

\usepackage{natbib}
 \bibpunct[, ]{(}{)}{,}{a}{}{,}%
 \def\bibfont{\small}%
\EquationsNumberedThrough    

\TheoremsNumberedThrough     
\ECRepeatTheorems  %

\MANUSCRIPTNO{MSOM-0001-2024.00}

\begin{document}


\RUNAUTHOR{Li et al.}
\RUNTITLE{Power Constrained Nonstationary Bandits with Habituation and Recovery Dynamics}

\TITLE{Power Constrained Nonstationary Bandits with Habituation and Recovery Dynamics}

\ARTICLEAUTHORS{%
\AUTHOR{Fengxu Li, Yonatan Mintz}
\AFF{Department of Industrial and Systems Engineering, University of Wisconsin-Madison, 
\EMAIL{\{fengxu.li,ymintz\}@wisc.edu}}

\AUTHOR{Stephanie M. Carpenter, Matthew P. Buman}
\AFF{College of Health Solutions, Arizona State University, 
\EMAIL{\{stephanie.m.carpenter,mbuman\}@asu.edu}}

} 

\ABSTRACT{
\textbf{Problem definition:} 
A common challenge for decision makers is selecting actions whose rewards are unknown and evolve over time based on prior policies. For instance, repeated use may reduce an action’s effectiveness (habituation), while inactivity may restore it (recovery). These nonstationarities are captured by the Reducing or Gaining Unknown Efficacy (ROGUE) bandit framework, which models real-world settings such as behavioral health interventions. While existing algorithms can compute sublinear regret policies to optimize these settings, they may not provide sufficient exploration due to overemphasis on exploitation, limiting the ability to estimate population-level effects.
This is a challenge of particular interest in micro-randomized trials (MRTs) that aid researchers in developing  just-in-time adaptive interventions that have population-level effects while still providing personalized recommendations to individuals. \textbf{Methodology/Results:}
We first develop ROGUE-TS, a Thompson Sampling algorithm tailored to the ROGUE framework, and provide theoretical guarantees of sublinear regret. We then introduce a probability clipping procedure to balance personalization and population-level learning, with quantified trade-off that balances regret and minimum exploration probability. Validation on two MRT datasets concerning physical activity promotion and bipolar disorder treatment shows that our methods both achieve lower regret than existing approaches and maintain high statistical power through the clipping procedure without significantly increasing regret. This enables reliable detection of treatment effects while accounting for individual behavioral dynamics.
\textbf{Managerial implications:}
For researchers designing MRTs, our framework offers practical guidance on balancing personalization with statistical validity. We demonstrate how prior data, such as from pilot studies, can be leveraged to inform adaptive intervention delivery. In cases where interventions differ in burden or risk, tailored strategies can be used to prioritize either participant safety or learning objectives. Importantly, we show that statistical power can still be achieved with reduced exploration, enabling the design of more efficient and participant-friendly trials.

}%





\maketitle

\section{Introduction}\label{sec:Intro}

Multi-armed bandits (MAB) are commonly used models for sequential decision making with partial information. They have been applied across various settings including healthcare  and online advertising \citep{villar2015multi,schwartz2017customer}. These models have not only been used for devising online decision making policies, but also as methods of data collection to inform population-wide decision making. Generally the models and methods proposed in these applications are meant to address the stationary MAB setting, when the reward distribution associated with each action does not change over time. However, the reward distributions in many real-world settings do change over time, necessitating the development of non-stationary MAB models. Such dynamic settings present a challenge because if policies do not properly account for non-stationarity the decision maker may experience linear regret  \citep{besbes2014stochastic,mintz2020nonstationary}. Moreover, since parameter estimation is more challenging in a non-stationary setting, it is unclear how decision makers could use parameter estimates from these models to inform population wide decisions. This is because methods that effectively optimize policies to non-stationary processes may not provide sufficient statistical power to the decision maker across the population to measure the effect size of their polices. Thus there is a critical need to develop methods that can both effectively optimize non-stationary models, and also provide data that can be used for population level statistical inference.

 This critical challenge is exemplified in applications such as personalized healthcare interventions and micro-randomized trials (MRTs) \citep{klasnja2015microrandomized,liu2023microrandomized}. Unlike traditional clinical trials, MRTs involve repeated randomization at multiple decision points per participant. At each point, an intervention is either delivered or withheld based on predefined probabilities, allowing researchers to observe how treatment effects vary in response to internal states (e.g., mood, fatigue) or contextual factors (e.g., location, time of day). The goal of these trials is to not only find the best treatment for each individual, but also use the gathered data to perform post-hoc hypothesis tests for population wide treatment effects. In these settings, since interventions are applied to a particular individual over time, the effectiveness of a particular intervention may decrease if the dosage (e.g., message delivery) is too high due to habituation while its effectiveness may increase if the dosage is lower due to perceived novelty. While MAB algorithms are well-suited for this setting, algorithms that are solely designed for regret minimization can over-adapt to early observations and prematurely discard certain treatments. This reduces the frequency with which some interventions are sampled, leading to biased or underpowered estimates of treatment effects \citep{bhatt2016adaptive}. This tension highlights the key challenge at the intersection of experimentation to optimize just-in-time adaptive interventions (JITAIs) and statistical inference, highlighting the need to design learning algorithms that remain responsive to individual dynamics while preserving the conditions necessary for drawing valid scientific conclusions. In this paper, we propose methods that produce non-stationary MAB policies that are able to negotiate this tradeoff both theoretically and computationally.

\subsection{Applications in MRTs}\label{sec:MotivatingApps}
Our proposed methods are motivated by applications in real world MRTs used for the optimization of behavioral interventions. To illustrate their utility, we provide examples of how our methods can apply to MRTs designed to optimize JITAIs for physical activity promotion.


Regular physical activity has many benefits for physical and mental well-being \citep{bull2020world}. Even modest increases in movement  (e.g., standing more often or walking short distances throughout the day) can lower the risk of cardiovascular disease, improve metabolic function, reduce symptoms of anxiety and depression, and enhance cognitive performance \citep{sattelmair2011dose,dunstan2012breaking,dempsey2016benefits}. Despite extensive evidence of its benefits, adherence to regular physical activity is difficult, particularly in settings where prolonged sitting predominates. Office work, commuting, and screen-based leisure contribute to daily routines that are often dominated by extended sedentary periods that not only reduce opportunities for movement but may also dampen the positive effects of otherwise adequate exercise \citep{owen2010too,rezende2014sedentary}.

In response, strategies encouraging brief, frequent bouts of movement, such as standing breaks or short walks, have gained attention as practical, scalable interventions. Mobile health (mHealth) platforms, including smartphone apps and wearable devices, show promise in delivering such interventions through real-time prompts. Despite these devices collecting rich, individualized physical activity data, many popular commercial trackers still provide users with one-size-fits-all, non-personalized activity goals and generic messages \citep{brickwood2019consumer}. This lack of personalization plausibly limits their effectiveness in sustaining user motivation and reducing sedentary behavior.
Conversely, studies indicate that context-aware, adaptive activity reminders, tailored to an individual’s momentary context and behavior, can lead to more significant reductions in sedentary time and improvements in physical activity levels \citep{fukuoka2011mped,klasnja2019}. 

While personalized, real-time prompts can effectively encourage physical activity, their success  may depend on timing, frequency, content, and individual differences \citep{heron2010ecological}. Repeated prompts may cause habituation, especially if they occur at inconvenient moments or lack sufficient variation. However, after a period without prompts, users’ sensitivity and motivation may partially recover. These patterns highlight the need for personalized interventions that can adapt in real-time to deliver the right treatment at the right time. MRTs provide a rigorous framework for testing such strategies, and the methods proposed in this paper are well posed to optimize timely, individualized physical activity support that accounts for habituation and recovery dynamics while still maintaining sufficient richness for statistical inference.

\subsection{Contributions}

In this paper, we address the challenge of developing methods for non-stationary MABs that effectively tradeoff strong regret guarantees with ensuring  statistical power. Our analysis is focused on the reducing or gaining unknown efficacy (ROGUE) multi-armed bandit framework \citep{mintz2020nonstationary}, which models nonstationary environments with frequently changing rewards that can accommodate behavioral effects such as habituation and recovery. Our contributions include:

\begin{enumerate}

\item We propose a novel Thompson sampling approach suited for ROGUE MAB models that we call ROGUE Thompson Sampling (ROGUE-TS). This is a randomized algorithm that can efficiently personalize in nonstationary environments. We prove that this approach is able to achieve regret of order $\mathcal{O}(T^{\frac{3}{4}})$, meaning it has an effective sublinear regret. This rate is significant because it is comparable to problem independent regret bounds for other algorithms designed for the ROGUE environment \citep{mintz2020nonstationary}. A randomized algorithm is key for forming the basis of treatment randomization in MRTs.

\item We propose a power-constrained framework within the ROGUE setting that ensures policies sufficiently explore all actions to achieve high statistical power. We do this by clipping the maximum and minimum probability an algorithm could choose any particular action while being implemented online. We determine the minimum clipping probability to enforce sufficient exploration such that we can achieve an effective tradeoff between regret and statistical power. Our theoretical analysis quantifies the trade-off between regret minimization and enforced exploration and shows that asymptotically as the number of participants increase we can achieve improved regret performance.

\item We validate our approach on two real-world datasets from MRTs. One MRT is focused on promoting physical activity and the other on bipolar disorder treatment. Our results show that our methods not only achieve better regret performance compared to state-of-the-art baselines but also maintain high statistical power without significantly increasing regret. This allows reliable detection of treatment effects while accounting for individual behavioral dynamics.
\end{enumerate}

\subsection{Literature review}\label{sec:LitReview}

Our research builds on three key areas of literature: multi-armed bandits (Section~\ref{sec:LitReviewBandits}), micro-randomized trial designs focused on causal inference (Section~\ref{sec:LitReviewMRT}), and the application of operations research methods in healthcare (Section~\ref{sec:LitReviewOR}).

\subsubsection{Multi-armed bandits}\label{sec:LitReviewBandits}
The multi-armed bandit problem has been extensively studied as a foundational framework for balancing exploration and exploitation, with many algorithms offering sublinear regret guarantees \citep{auer2002finite, auer2002nonstochastic, l2o}. In practice, bandit algorithms have been applied to support decision-making in diverse domains, including personalized interventions \citep{tewari2017ads, lattimore2016causal}, resource scheduling \citep{gittins1979bandit, lai1985asymptotically}, and experimentation design \citep{villar2015multi, bastani2020online}, all of which are highly relevant to healthcare applications. However, many healthcare settings involve nonstationary environments in which patient states and treatment effects evolve over time \citep{wood2008empirical}. This nonstationarity is not unique to healthcare and arises in various other domains as well, motivating the development of bandit algorithms that can adapt to changing reward distributions. Common approaches for handling nonstationarity include sliding-window estimation, exponential discounting, change-point detection, and explicit modeling of dynamic environments \citep{besbes2014stochastic, garivier2008upper, pmlr-v75-luo18a, mintz2020nonstationary,he2024non}.

Although MAB algorithms are well-suited to online optimization, most are designed primarily to minimize regret rather than for statistical inference, an equally critical goal in healthcare research \citep{glass2013causal}. Adaptive treatment assignments can introduce selection bias that threatens valid hypothesis testing \citep{villar2015multi}. Notably, \cite{yao2021power} proposed a minimum exploration mechanism within contextual bandits to balance personalization with valid inference, demonstrating promising results. However, such approaches remain scarce in nonstationary settings, revealing a significant gap in the literature.
Most existing methods estimate treatment effects by embedding individual variation within a static, population-level model. In contrast, our approach adopts a different perspective: it explicitly separates individual dynamics from population-level treatment effects, allowing for inference that captures nonstationary patterns, such as those frequently observed in MRTs. To address this critical gap, we develop a constrained bandit framework that incorporates statistical power requirements in the presence of nonstationarity.

\subsubsection{Micro-randomized trials}\label{sec:LitReviewMRT}
A related yet distinct line of work focuses on MRTs, a widely used experimental design in mHealth research to evaluate the proximal causal effects of interventions delivered in real time \citep{klasnja2015microrandomized, Liao2016}. Classical MRTs use fixed, pre-specified randomization probabilities at each decision point to ensure sufficient exploration and unbiased estimation of treatment effects \citep{klasnja2015microrandomized, Qian2022}. This design has been effectively applied across a range of mHealth domains, including physical activity \citep{klasnja2019}, smoking cessation \citep{battalio2021sense2stop}, mental health \citep{necamp2020assessing}, and alcohol use self-monitoring \citep{carpenter2023self}. However, fixed randomization strategies do not account for dynamic user states or heterogeneous treatment responsiveness, which may lead to suboptimal intervention delivery  \citep{nahum2018just, Dempsey2020,pulick2025adaptive,pulick2025idiographic}. To improve personalization while maintaining  statistical power, researchers have proposed more adaptive randomization strategies. These include context-adaptive designs that allow for estimation of effect moderation across different contextual strata \citep{Boruvka2018}, and stratified MRTs that dynamically adjust treatment probabilities based on predicted user risk or receptivity \citep{Dempsey2020}. Our approach is conceptually aligned with contextualized randomization strategies but differs in implementation. Rather than relying on predefined rules to update randomization probabilities, we use a constrained MAB framework that adaptively adjusts treatment assignments based on observations. By explicitly modeling user states, our method aims to deliver more personalized interventions while preserving the ability to draw valid causal inferences.

\subsubsection{Operations research applications in healthcare}\label{sec:LitReviewOR}

Another related stream of research illustrates how operations research (OR) methods can support healthcare decision-making. These applications span disease screening \citep{lee2019optimal, ayer2016heterogeneity,li2022optimal}, treatment regimens \citep{alagoz2004optimal, he2023model,olfat2020flexible}, hospital operations \citep{fairley2019improving, helm2014design}, personalized interventions \citep{aswani2019behavioral, baek2025policy}, and public health planning \citep{adams2023planning, brandeau2003resource}. Our work falls under personalized interventions and shares similarities with the approach of \cite{aswani2019behavioral}, which modeled patient behavior in a weight loss program using feedback to update estimates of individual states. Like our study, they  leverage adaptive decision-making based on evolving patient information. In contrast to their work and much of the existing OR literature, which focuses on optimizing outcomes, our contribution lies in examining how causal inference methods can be integrated with optimization under the partial information settings common in healthcare decision making. We do this by studying the tradeoff between obtaining statistically rich data, which will result in better population wide outcomes, and optimization of personal outcomes of participants.

\section{Problem setting and preliminaries}
We now formalize the problem setting and introduce the key technical assumptions that underpin our  analysis. These elements are drawn partially from the ROGUE framework \citep{mintz2020nonstationary}.

\subsection{ROGUE bandit setting}
In the ROGUE model, the decision maker is given a fixed finite set of arms $\mathcal{A}$ over a known time horizon $T$.  
At each round $t \in \{1, \dots, T\}$, the decision maker selects an action $\pi_t \in \mathcal{A}$ based on the history $H_t$, where the policy $\Pi = \{\pi_t(\cdot)\}_{t=1}^T$ is a sequence of decision rules
$\pi_t: H_t \to \mathcal{A},$ mapping the history $H_t$ into an action.
Let $r_{a,s}$ denote the reward received from arm $a$ at time $s$. The history $H_t$ is the cumulative record of past actions and rewards before time $t$: $H_t = \{(\pi_1, r_{\pi_1, 1}), (\pi_2, r_{\pi_2, 2}), \ldots, (\pi_{t-1}, r_{\pi_{t-1}, t-1})\}$. Each action $a\in \mathcal{A}$ at time $t$, if selected, yields a stochastic reward $r_{a, t}$ drawn from a sub-Gaussian distribution $\mathbb{P}_{\theta_a, x_{a,t}}$ with expectation
$\mathbb{E}[r_{a,t}] = g(\theta_a, x_{a,t}),$ where $g$ is a bounded function.
The parameters $\theta_a \in \Theta$ for each arm $a \in \mathcal{A}$ are unknown constants but lie in a known compact convex set $\Theta$.
Each action $a$ also has an associated state $x_{a,t} \in \mathcal{X}$ that evolves over time according to the known dynamics:
$x_{a,t+1} = h_a(x_{a,t}, \pi_{a,t}),$
where $\pi_{a,t} = \mathbf{1}[\pi_t = a]$ indicates whether arm $a$ was selected at time $t$, and $h : \mathcal{X} \times \mathbb{B} \to \mathcal{X}$ is a known function. The state space $\mathcal{X}$ is compact and convex, and initial states $x_{a,0}$ are unknown but lie in $\mathcal{X}$. The decision maker aims to minimize their dynamic regret, defined as the difference between the cumulative expected reward of the actions played by an oracle with knowledge of the true system parameters  and the reward obtained by the chosen policy $\Pi$:
$R_{\Pi}(T) =  \mathbb{E}[\sum_{t=1}^T r_{\pi_t^*, t} - r_{\pi_t, t}],$
where $\pi_t^*$ denotes the arm played by the oracle at time $t$.
The objective is to design an admissible policy $\Pi$ that minimizes this expected regret. Alternatively, this problem can be thought of as finding a policy that maximizes the expected reward.

\subsection{Assumptions}
To analyze our methods we require the following set of technical assumptions:

\begin{assumption}
    The rewards $r_{a,t}$ are conditionally independent given $x_{a,0}$, $\theta_a$ (or equivalently the complete sequence of $x_{a,t}$, $\pi_t$ and $\theta_a$).
    \label{dependenceAssum}
    \end{assumption}
    
    For two distinct time points $t$ and $t'$, $r_{a,t}|\{\theta,x_{a,t}\}$ does not depend on $r_{a,t'}|\{\theta,x_{a,t'}\}$. This mirrors the independence assumption in stationary MAB but relaxes it to this non-stationary setting.

    \begin{assumption}
    \label{logconcave}
    The reward distribution $P_{\theta,x}$ has a log-concave probability density function (p.d.f.) $p(r|\theta, x)$ for all $x \in \mathcal{X}$ and $\theta \in \Theta$.
    \end{assumption}
This assumption provides regularity for the reward distributions and is met by many common distributions (e.g., Gaussian and Bernoulli).

Next, we define a function $ f(\cdot) $ to be $ L $-Lipschitz continuous if for all $ x_1, x_2 $ in the domain of $ f $, it holds that
$|f(x_1) - f(x_2)| \leq L \|x_1 - x_2\|_2.$
Similarly, $ f(\cdot) $ is said to be $ L $-bi-Lipschitz continuous if for all $ x_1, x_2 $ in the domain of $ f $,
$\frac{1}{L} \|x_1 - x_2\|_2 \leq |f(x_1) - f(x_2)| \leq L \|x_1 - x_2\|_2.$
We say that $f$ is locally $ L $-Lipschitz (respectively, locally $ L $-bi-Lipschitz) on a compact set $\mathcal{S}$ if the Lipschitz (respectively, bi-Lipschitz) condition holds for all pairs $ x_1, x_2 \in \mathcal{S} $.
The next assumption concerns the stability of the distributions with respect to various parameters.

    \begin{assumption}
    \label{lipschitz}
    The log-likelihood ratio $\ell(r; \theta', x', \theta, x) = \log \frac{p(r|\theta', x')}{p(r|\theta, x)}$ associated with the distribution family $\mathbb{P}_{\theta,x}$ is locally $L_f$-Lipschitz continuous with respect to $x, \theta$ on the compact set $\mathcal{X} \times \Theta$ for all values of $\theta', x' \in \mathcal{X} \times \Theta$, and $g$ is locally $L_g$-bi-Lipschitz continuous with respect to $x, \theta$ on the compact set $\mathcal{X} \times \Theta$.
    \end{assumption}
This condition ensures that small changes in the parameters lead to small changes in the log-likelihood ratio, implying stability in the distribution with respect to parameters.
    
    \begin{assumption}
    \label{rewardAssum}
    The reward distribution $\mathbb{P}_{\theta,x}$ for all $\theta \in \Theta$ and $x \in \mathcal{X}$ is sub-Gaussian with parameter $\sigma$, and either $p(r|\theta, x)$ has a finite support or $\ell(r; \theta', x', \theta, x)$ is locally $L_p$-Lipschitz with respect to $r$.
    \end{assumption}
This assumption ensures that sample averages will closely approximate their expected values.
    
    Finally, we set conditions on the dynamics governing the state of each action:
        
    \begin{assumption}
    The dynamic transition function $h$ is $L_h$-Lipschitz continuous such that $L_h \leq 1$.
    \end{assumption}
This assumption ensures that changes in the state of each action are gradual, implying stability in the system’s dynamics.

\subsection{Hypothesis testing in the ROGUE setting}\label{modelExtension}

Building upon the original ROGUE model and its underlying assumptions, we extend the framework to handle multiple users and incorporate MRT-style inference through formal hypothesis testing. Our extension considers a population of $N$ users indexed by $i \in \{1, \dots, N\}$, each exhibiting individualized state dynamics and reward responses. At each time step $t$, the decision maker selects an action $a_{i,t} \in \mathcal{A}$ for user $i$. The state associated with user $i$ and arm $a$ at time $t$, denoted $x_{i,a,t}$, evolves according to a known dynamics function $h_{i,a}$, such that
$x_{i,a,t+1} = h_{i,a}(x_{i,a,t}, \pi_{i,a,t}),$
where $\pi_{i,a,t} = \mathbf{1}[a_{i,t} = a]$ indicates whether arm $a$ was selected for user $i$ at time $t$. Upon selecting arm $a$, the decision maker observes a stochastic reward $r_{i,a,t}$ with expected value
$\mathbb{E}[r_{i,a,t}] = g(\theta_a, x_{i,a,t}),$
where $\theta_a \in \Theta \subset \mathbb{R}$ is a fixed, arm-specific treatment effect shared across users, and $x_{i,a,t}$ is user-specific.

While this general formulation supports  individualized modeling, inference becomes computationally challenging and potentially non-identifiable in high-dimensional state spaces. To address this, we restrict our analysis to a linear dynamical system with scalar state (i.e.,$x_{i,a,t}\in \mathbb{R}$) and additive reward structure. This simplification allows for tractable inference and aligns with MRT-style hypothesis testing frameworks. We assume the expected reward decomposes linearly as  $\mathbb{E}[r_{i,a,t}] = g(\theta_a, x_{i,a,t}) = C x_{i,a,t} + D \theta_a,$ 
where $C \in \mathbb{R}$ is a known coefficient capturing the contribution of the state to the reward and $D \in \mathbb{R}$ is a known scalar determining the magnitude of the treatment effect.
Designating one arm $a' \in \mathcal{A}$ as the baseline (i.e., a \emph{do-nothing} action), we test whether any alternative arm $a \ne a'$ yields a statistically significant improvement. The hypothesis test is conducted at level $\alpha$ with target power $1 - \beta$:$   H_0: \theta_a = \theta_{a'} \text{ vs. }  H_1: \theta_a > \theta_{a'}.$ The decision maker’s objective is to design a policy that achieves the desired statistical power while minimizing dynamic regret.

\section{Thompson sampling in the ROGUE setting}

Thompson sampling (TS) has been shown to achieve sublinear regret in the stationary MAB setting \citep{l2o}. However, its performance degrades in nonstationary environments, where reward distributions may evolve faster than TS can adapt. In such cases, TS may fail to track shifting reward dynamics, resulting in increased regret \citep{nmabLiu}. To overcome this challenge, we propose ROGUE-TS, an extension of TS within the ROGUE framework that explicitly models nonstationary reward dynamics. ROGUE-TS leverages Bayesian filtering techniques to update posteriors over time, enabling adaptive intervention strategies in dynamic settings. In this section, we present the algorithm and analyze its regret guarantees under nonstationary conditions. Note that in this section we provide an overview of the main results and sketch of the proofs, all detailed proofs for propositions, lemmas, and theorems in this section are provided in Appendix~\ref{ap:proof}.

\subsection{Algorithm description}\label{algoTS}

Pseudocode for ROGUE-TS is provided in Algorithm~\ref{alg:rogue-ts}, and the algorithm is written for $T$ time periods labeled $t = 1, \dots, T$. At each time step, the policy selects an action from a finite set of arms $\mathcal{A}$ based on samples drawn from the posterior distribution over true parameters associated with each arm. The algorithm maintains a posterior distribution of the true parameters $(\theta_a, x_{a,t})$ for each arm $a$, which is updated over time as rewards are observed. During the initialization phase, the policy begins with prior beliefs over each arm’s parameters. Then, at each time step, the algorithm proceeds by sampling a candidate pair $(\tilde{\theta}_{a}, \tilde{x}_{a,t})$ from the posterior for each arm $a\in \mathcal{A}$. Using the reward function $g$, the algorithm computes a sampled reward $\tilde{r}_{a,t} = g(\tilde{\theta}_{a}, \tilde{x}_{a,t})$, and the arm with the highest sampled reward is selected and played. After receiving a reward, ROGUE-TS updates the posterior distributions in two stages: correction and prediction. In the correction step, only the posterior of the played arm is updated by incorporating the newly observed reward through Bayesian conditioning. In the prediction step, the beliefs for all arms are propagated forward according to the system dynamics $h$. The posteriors of the unplayed arms are updated solely via prediction. This procedure performs exact Bayesian inference, meaning the posterior distributions are computed analytically without approximation. Here, $\delta(\cdot)$ refers to the Dirac delta function, which enforces a deterministic constraint by assigning all probability mass to a single point.

\SingleSpacedXII
\begin{algorithm}[H]
    \caption{Genral ROGUE-TS}
    \label{alg:rogue-ts}
   \begin{algorithmic}[1]
      \Require Set of arms $\mathcal{A}$; initial prior distributions $\mathbb{Q}_a^{(1)}$ over $(\theta_a, x_{a,1})$ for each $a \in \mathcal{A}$
     
     \For{$t = 1$ to $T$}
        \For{each $a \in \mathcal{A}$}
            \State Sample $(\tilde{\theta}_{a}, \tilde{x}_{a,t}) \sim \mathbb{Q}_a^{(t)}$
            \State Compute sampled reward $\tilde{r}_{a,t} = g(\tilde{\theta}_{a}, \tilde{x}_{a,t})$
        \EndFor
        \State Select arm $a_t = \arg\max_{a \in \mathcal{A}} \tilde{r}_{a,t}$
        \State Play arm $a_t$ and observe reward $r_{a_t,t}$
        \Statex \quad \textbf{Update posterior for selected arm $a_t$ (correction + prediction):}
       
      \State $\mathbb{Q}_{a_t}^{(t+1)}(\theta_{a_t}, x_{a_t,t+1}) \propto \mathbb{Q}_{a_t}^{(t)}(\theta_{a_t}, x_{a_t,t}) \cdot p(r_{a_t,t} \mid \theta_{a_t}, x_{a_t,t}) \cdot \delta\big(x_{a_t,t+1} - h_{a_t}(x_{a_t,t}, 1)\big)$
         
        \For{each $a \in \mathcal{A} \setminus \{a_t\}$}
            \Statex \hspace{\algorithmicindent}
            \quad\textbf{Prediction-only update:}
            
        \State   $ \mathbb{Q}_a^{(t+1)}(\theta_a, x_{a,t+1}) = \int \mathbb{Q}_a^{(t)}(\theta_a, x_{a,t}) \cdot \delta\big(x_{a,t+1} - h_a(x_{a,t}, 0)\big) \, dx_{a,t}  $   
        \EndFor
     \EndFor
    \end{algorithmic}
\end{algorithm}
\OneAndAHalfSpacedXII
In practice, ROGUE-TS uses the Kalman Filter when the underlying model exhibits linear dynamics with Gaussian noise, enabling efficient, closed-form Bayesian updates \citep{zarchan2000kalman}. We also provide the implementation of this setting in Appendix~\ref{app:kalman-rogue}. In more complex settings involving nonlinear dynamics or non-Gaussian noise, we still aim to perform exact Bayesian updates through more specific or model-dependent calculations. However, when exact computation becomes intractable, approximation methods such as particle filtering \citep{djuric2003particle}, which estimates the posterior using a weighted set of samples can be applied.

\subsection{Regret analysis}
Since ROGUE-TS is a Bayesian algorithm, in this section we analyze its Bayse regret, where the expectation in $R_{\Pi}(T)$ is taken over the prior distribution of the unknown initial parameters as well as the randomness in both the rewards and the algorithm’s action selection.
Our key result for  ROGUE-TS is to show that it achieves sublinear dynamic regret.

\begin{theorem}\label{thmReg}
    For any $T\in \mathbb{N}$: $R_\Pi(T) \leq  \frac{8}{3}L_g^2 |\mathcal{A}|^{\frac{1}{4}}T^{\frac{3}{4}}\sqrt{ 2\sigma c_f(d_x,d_\theta) + 4L_p\sigma^2\sqrt{\log(T)}} +2C_g\min\{T,|\mathcal{A}|\}.$
\end{theorem}

To establish our result, we build on a key structural insight: a regret decomposition technique introduced by \cite{l2o}, which relates the expected regret of TS to that of a UCB-based policy. We then construct high-probability bounds for the evolving reward distributions using trajectory-based Kullback–Leibler divergence \citep{mintz2020nonstationary}. These bounds give rise to UCB-style indices that serve as analytical surrogates for evaluating the performance of TS.

Following \cite{l2o}, a confidence bound sequence is a pair of collections of functions $ (L, U) = \{(L_t, U_t) \mid t \in \mathbb{N} \} $, where each $ L_t $ and $ U_t $ map the history $ H_t $ to functions $ L_t(H_t), U_t(H_t): \mathcal{A} \to \mathbb{R} $, representing lower and upper confidence bounds for the expected reward of each action at time $ t $.
For notational simplicity, we write $ L_t(a) $ and $ U_t(a) $ to denote the lower and upper confidence bounds assigned to action $ a \in \mathcal{A} $ at time $ t $, based on $ H_t $.

\begin{proposition}\label{decom}
For any upper confidence bound sequence $U = \{U_t \mid t \in \mathbb{N}\}$, the regret can be decomposed as
$    \mathbb{E}(r_{\pi^*_t,t} - r_{\pi_t,t}) = \mathbb{E}(r_{\pi^*_t,t} - U_t(\pi^*_t)) + \mathbb{E}(U_t(\pi_t) - r_{\pi_t,t}),$
at each time step $t$.
\end{proposition}

This decomposition expresses the regret as the sum of two interpretable terms: the \emph{underestimation error} for the optimal action and the \emph{overestimation error} for the chosen action. These terms provide a natural interface for UCB-style analysis, even when actions are selected via TS. To quantify the difference between the true and estimated parameters, we use the notion of trajectory Kullback–Leibler (KL) divergence, introduced in \cite{mintz2020nonstationary}. This divergence measures the cumulative discrepancy between two reward-generating processes along the trajectory induced by a policy, which is especially relevant in nonstationary settings where states evolve over time.

\begin{definition}
\textit{Given an input action sequence $\pi_1^T$, let $\mathcal{T}_a$ denote the set of time steps up to time $T$ at which action $a \in \mathcal{A}$ was selected by $\pi_1^T$, that is, $u_{a,t} = 1$ for all $t \in \mathcal{T}_a$. The \emph{trajectory KL divergence} between two parameter-state pairs $(\theta_a, x_{a,0})$ and $(\theta'_a, x'_{a,0})$ in $\Theta \times \mathcal{X}$ is defined as:  $      D_{a,\pi_1^T}(\theta_a, x_{a,0} \,\|\, \theta'_a, x'_{a,0}) 
        = \sum\limits_{t \in \mathcal{T}_a} D_{\mathrm{KL}}\big( \mathbb{P}_{\theta_a, x_{a,t}} \,\|\, \mathbb{P}_{\theta'_a, x'_{a,t}} \big) = \sum\limits_{t \in \mathcal{T}_a} D_{\mathrm{KL}}\big( \mathbb{P}_{\theta_a, h_a^t(x_{a,0})} \,\|\, \mathbb{P}_{\theta'_a, h_a^t(x'_{a,0})} \big),$ where $h_a^t$ denotes the $t$-step composition of the dynamics function $h_a$ applied to the initial state. The notation $\mathbb{P}_{\theta,x}$ is the reward distribution induced by parameters $(\theta, x)$, and $D_{\mathrm{KL}}$ is the standard KL divergence.}
\end{definition}

Let $(\hat{\theta}_a, \hat{x}_{a,0})$ be the maximum likelihood estimate (MLE) of the parameters based on observed rewards, and $(\theta_a^*, x_{a,0}^*)$be the true parameters. The MLE is used to construct confidence intervals. The following concentration inequality, bounds the trajectory KL divergence between the true and estimated parameters with high probability.

\begin{theorem}[Concentration Bound, Corollary 1 in \cite{mintz2020nonstationary}]
\label{concentration}
    For $\alpha \in (0,1)$, with probability at least $1-\alpha$ we have
   $ \frac{1}{n(\mathcal{T}_a)}D_{a,\pi_1^{T}}(\theta^*_a,x^*_{a,0}||\hat{\theta}_a,\hat{x}_{a,0}) \leq B(\alpha)\sqrt{\frac{\log(1/\alpha)}{n(\mathcal{T}_a)}},$
where $B(\alpha) = \frac{c_f(d_x,d_\theta)}{\sqrt{\log(1/\alpha)}}+L_p\sigma\sqrt{2},$ with
      $  c_f(d_x,d_\theta) = 8L_f\diam(\mathcal{X})\sqrt{\pi} + 48\sqrt{2}(2)^\frac{1}{d_x+d_\theta}L_f\diam(\mathcal{X}\times\Theta)\sqrt{\pi(d_x+d_\theta)}  $
is a constant and $\diam(\mathcal{X})$ represents the diameter of the set $\mathcal{X}$, which is defined as the largest possible distance between any two points in the set.
\end{theorem}

Define
$g_{\max} = \max\{ g(\theta, x) \mid (\theta, x) \in \Theta \times \mathcal{X} \}$, 
$g_{\min} = \min\{ g(\theta, x) \mid (\theta, x) \in \Theta \times \mathcal{X} \}$, 
and let the reward range constant be $C_g = g_{\max} - g_{\min}$, which will be used to bound the maximum regret in one arm pull. We denote $\mathcal{T}_{a,t} \subset \{1,2,\dots, t\}$ as the set of time steps up to time $t$ during which action $a$ was selected, and let $n(\mathcal{T}_{a,t}) = |\mathcal{T}_{a,t}|$. Similarly, the notation $D_{a,\pi_1^t}(\theta_a, x_{a,0} \,\|\, \theta'_a, x'_{a,0})$ refers to the trajectory KL divergence computed only up to time $t$, rather than the entire horizon $T$. The following proposition gives a lower bound of this trajectory KL divergence.

\begin{proposition}
\label{past-reward-kl-divergence}
For any $ (\theta_a, x_{a,0}), (\theta_a', x_{a,0}') \in \Theta \times \mathcal{X} $, the average trajectory KL divergence between the reward distributions induced by the two pairs over the time steps $ \mathcal{T}_{a,t} $ satisfies the following lower bound:
 $   \frac{1}{n(\mathcal{T}_{a,t})} D_{a,\pi_1^t}(\theta_a, x_{a,0} \,\|\, \theta_a', x_{a,0}') 
\geq \frac{1}{2\sigma n(\mathcal{T}_{a,t})} \sum\limits_{t'\in \mathcal{T}_{a,t}}\left(g(\theta,h_a^{t'}(x_0)) - g({\theta}'_a,h_a^{t'}({x}'_{a,0}))\right)^2.$
\end{proposition}

The result follows from Assumption \ref{rewardAssum}, which implies that the rewards are sub-Gaussian with parameter $\sigma$. This allows us to invoke the fact that exponential family distributions attain maximum entropy under moment constraints, leading to the stated bound \citep{bickel2015mathematical}.

We now define a function that characterizes the worst-case deviation in expected reward at time $ t $ under a constraint on the cumulative reward discrepancy. This function enables the construction of UCB-style confidence intervals based on the divergence bound established earlier:
   $\psi_{a,t}(\gamma) = \max \bigg\{  \left| g(\theta,h_a^t(x_0)) - g(\hat{\theta}_a,h_a^t(\hat{x}_{a,0})) \right| : 
     \sum\limits_{t'\in \mathcal{T}_{a,t}} \frac{\left(g(\theta,h_a^{t'}(x_0)) - g(\hat{\theta}_a,h_a^{t'}(\hat{x}_{a,0}))\right)^2 }{2\sigma n(\mathcal{T}_{a,t})} \leq \gamma \bigg\},$
where 
$\gamma$ is a parameter controlling the confidence level. This function captures the worst case reward deviation for action 
$a$ at time $t$, subject to the lower bound of the KL divergence constraint and can be bounded as follows:
\begin{lemma}
\label{lemma1}
If $\psi_{a,t}(\gamma)$ is defined as above, then
 $   \psi_{a,t}(\gamma) \leq L_g^2\sqrt{2\sigma \gamma}.
    $\label{boundsOnPsi}
\end{lemma}
This follows by recursively applying the Lipschitz continuity of $g$ and $h$, and observing that the difference $\left|g(\theta, h_a^t(x_0)) - g(\hat{\theta}_a, h_a^t(\hat{x}_{a,0}))\right|$
does not increase as $t$ decreases, yielding a uniform lower bound across $\mathcal{T}_{a,t}$. For simplicity, we also provide an upper bound for regret at each time $t$.
\begin{proposition} \label{prop2}
    For any upper confidence bound sequence $U = \{U_t|t \in N\}$ and lower confidence bound sequence $L = \{L_t|t \in N\}$, the regret has an upper bound as the following form:
    $\mathbb{E}(r_{\pi^*_t,t}-r_{\pi_t,t})\leq (U_t-L_t)(\pi_t) + \sum\limits_{a \in \mathcal{A}}C_g {P}\left(  \{g(\theta_{a}^* , h_a^{t}(x_{a,0}^*) )\notin [L_t(a),U_t(a)] \}\right)$
    at each time step $t$.
\end{proposition}

Using the decomposition in Proposition~\ref{decom}, we bound each term by considering whether the expected reward lies within the confidence interval. In both cases, the regret is at most the interval width plus a term bounded by 
$C_g$ times the probability that the expected reward falls outside the interval. Summing both terms gives the result. We construct an interval $[L_t(a), U_t(a)]$ around the predicted reward $g(\hat{\theta}_a, \hat{x}_{a,t})$. If an arm has never been selected ($n(\mathcal{T}_{a,t})=0$), we set $L_t(a)=g_{\min}$ and $U_t(a)=g_{\max}$ as the most conservative choice. Otherwise, the interval is refined using the control factor $\gamma_{a,t}$ through $\psi_{a,t}(\gamma_{a,t})$, ensuring the bounds hold with high probability. Formally, $\gamma_{a,t} = \frac{c_f(d_x,d_\theta) + 2L_p\sigma\sqrt{\log(T)}}{\sqrt{n(\mathcal{T}_{a,t})}}, \; U_t(a) = g_{\max}\mathbf{1}[n(\mathcal{T}_{a,t})=0] + \min \{g_{\max}, \, g(\hat{\theta}_a, \hat{x}_{a,t}) + \psi_{a,t}(\gamma_{a,t})\}\mathbf{1}[n(\mathcal{T}_{a,t}) \geq 1], \; L_t(a) = g_{\min}\mathbf{1}[n(\mathcal{T}_{a,t})=0] + \max \{g_{\min}, \, g(\hat{\theta}_a, \hat{x}_{a,t}) - \psi_{a,t}(\gamma_{a,t})\}\mathbf{1}[n(\mathcal{T}_{a,t}) \geq 1].$ Thus, the true parameters lie within these confidence intervals with high probability.

\begin{lemma}
\label{lemma2}
    If $ U_t{(a)} $ and $ L_t{(a)} $ are defined as above, then we have the following result:
     $   \sum\limits_{t=1}^T{P}\left(\{ g(\theta_{a}^*, h_a^{t}(x_{a,0}^*)  )\notin [L_t(a),U_t(a)]  \}\right)\leq 1/T
       $ 
\end{lemma}
This follows by applying the concentration inequality from Theorem~\ref{concentration} with $\alpha = 1/T^2$, together with the KL divergence lower bound established in Proposition~\ref{past-reward-kl-divergence}.

Now we provide a sketch of the Proof of Theorem \ref{thmReg}. We first apply Proposition~\ref{prop2} to obtain an upper bound on the regret, which consists of two terms. The first term, corresponding to the difference between $U_t(a)$ and  $L_t(a)$, is bounded using Lemma~\ref{lemma1}. We then upper bound the sum using an integral approximation followed by Jensen’s inequality to show that this term contributes a sublinear amount $O(T^{3/4})$ to the final regret bound.
For the second term, involving the probability
$\sum\limits_{t=1}^T{P}\left(  \{g(\theta_{a}^* , h_a^{t}(x_{a,0}^*) )\notin [L_t(a),U_t(a)] \}\right),$
we use Lemma~\ref{lemma2} to bound this event by a constant depending on $\min\{T, |\mathcal{A}|\}$. Summing these results yields the desired conclusion.

\section{Inclusion of power constraints}\label{PCModel}

In this section, we examine an application of our ROGUE-TS algorithm in the context of MRTs, where a randomized assignment strategy can be effectively employed. While the previous section allowed for general modeling assumptions to support algorithmic flexibility, enabling rigorous statistical inference requires identifiability, which in turn calls for more structured model assumptions.

Specifically, we focus on a particular model introduced in Section~\ref{modelExtension}, which ensures the well-posedness of parameter estimation. In this setting, the dynamics for each user evolve according to linear dynamics
$x_{t+1} = A x_{i,a,t} + B \pi_{i,a,t} + K, $ for $t = 0, 1, \dots, T-1,$
and the expected reward is given by
$\mathbb{E}r_{i,a,t} = g(\theta_a, x_{i,a,t}) = C x_{i,a,t} + D \theta_a.$ Recall that our goal is to test the hypotheses  $H_0: \theta_a = \theta_{a'}$  v.s. $H_1: \theta_a > \theta_{a'}$ between a given baseline arm $a'$ and other treatment arms $a$. To ensure identifiability and statistical independence in this extended model, we impose the following assumption:

\begin{assumption}
    \label{extendDependence}
    We assume that $|A| \neq 1$ or $0$ to avoid degenerate or non-informative state trajectories (e.g., constant or explosive dynamics), which would compromise identifiability. Additionally, we assume that the reward processes are independent across users; specifically, for any users $i \neq i'$, the reward $r_{i,a,t}$ is independent of $r_{i',a',t'}$ for all $a, a' \in \mathcal{A}$ and $t, t' \in \{1, \dots, T\}$.
\end{assumption}
Furthermore, in this section, for simplicity, we abuse the notation $\sigma^2$ to denote the variance of each $r_{i,a,t}$, and assume their sub-Gaussian parameter is $\overline{\sigma}$, which will be used solely in the proof.

Although ROGUE-TS is inherently randomized, it may increasingly favor exploitation over exploration as learning progresses, potentially leading to imbalanced treatment assignments. This motivates the introduction of a clipping probability mechanism, designed to enforce power constraints and guarantee adequate data collection across all intervention arms for reliable estimation. To guide the design of this mechanism, we first derive a least squares estimator for each user, then aggregate these into a pooled estimator across the entire population. We analyze how to control the variance of these estimators and explain why minimizing this variance is crucial for ensuring valid and powerful hypothesis testing.
Note that in this section we provide a discussion of the main results and a sketch of the proofs, all detailed proofs for propositions, lemmas, and theorems in this section are provided in Appendix~\ref{ap:proof}.

\subsection{Estimator of treatment effect}

To investigate how to set the clipping probabilities effectively, we begin with the estimator of the shared treatment effect $\theta_a$ for each arm $a$. To do this, we first construct an individual least square estimator, denoted $\hat{\theta}_{i,a}$, for each user $i$ based on their observed trajectory. Here, a trajectory refers to the sequence of observations collected from user $i$ over time under the intervention of arm $a$. Estimating parameters separately for each user allows us to incorporate individual differences and capture variability in their response. Then we form a single estimate of $\theta_a$ that pools information across all users, we define the aggregated estimator as the average of the individual user estimators: $\hat{\theta}_a = \frac{1}{N} \sum_{i=1}^N \hat{\theta}_{i,a}.$ Our goal is to show that this estimator is asymptotically normally distributed.

\begin{theorem}
\label{individual-MLE}
Let user $i$ receive arm $a$ at time steps $\mathcal{T}_{i,a} \subseteq \{1, \dots, T\}$ over the time horizon $T$. Let the joint parameter vector be $z_{i,a} = [x_{i,a,0}, \theta_a]^\top \in \mathbb{R}^2$. Define the least square estimator as the solution to
$(\hat{x}_{i,a,0},\, \hat{\theta}_{i,a}) = \argmin\limits_{x_{i,a,0},\, \theta_a} \Bigg\{  \sum\limits_{t \in \mathcal{T}_{i,a}} \left(r_{i,a,t}-g(x_{i,a,t},\theta_a)\right)^2 \; : 
 x_{i,a,t+1} = h_a(x_{i,a,t}, \pi_{i,a,t}), \text{ for } t \in \{0, \dots, T\} \Bigg\}.$
Then, the estimator $\hat{\theta}_{i,a}$ follows a sub-Gaussian distribution with mean $\theta_a^*$ and variance $\left[\mathcal{I}_{i,a}(z_{i,a}^*)^{-1}\right]_{2,2}$,
where $\theta_a^*$ denotes the true parameter value and $z_{i,a}^*$ denote the true parameter vector, and $\mathcal{I}_{i,a}(z_{i,a}^*)$ is the design matrix defined as
$\mathcal{I}_{i,a}(z_{i,a}^*) = \frac{1}{\sigma^2} \sum_{t \in \mathcal{T}_{i,a}}
\begin{bmatrix}
(C A^{t})^2 & C A^{t} D \\
C A^{t} D & D^2
\end{bmatrix}.$
In particular, the variance of $\hat{\theta}_{i,a}$ is
$\var(\hat{\theta}_{i,a}) = {\sigma^2}/\left(
   D^2 |\mathcal{T}_{i,a}| 
    - D^2\dfrac{\left( \sum_{t \in \mathcal{T}_{i,a}} A^t \right)^2}{\sum_{t \in \mathcal{T}_{i,a}} (A^t)^2}
  \right).$
Moreover, as the number of users $N \to \infty$, the pooled estimator $\hat{\theta}_a = \frac{1}{N} \sum_{i=1}^N \hat{\theta}_{i,a}$ is asymptotically normal. For large $N$, its distribution can be approximated by:
        $\hat{\theta}_a \approx \mathcal{N} \left(
        \theta_a^*,\;
        \left( \frac{1}{N^2} \sum_{i=1}^N \var(\hat{\theta}_{i,a})\right)^{1/2}
        \right).$
\end{theorem}

To prove the theorem, we begin by unrolling the state dynamics to express the state $x_{i,a,t}$ in terms of the initial state $x_{i,a,0}$, known parameters, and past actions. Substituting this into the reward model gives an expression where the observed reward depends linearly on the unknowns $x_{i,a,0}$ and $\theta_a$, plus Gaussian noise.
We jointly estimate $x_{i,a,0}$ and $\theta_a$, and show that the resulting estimator $\hat{z}_{i,a}$ is unbiased and sub-Gaussian distributed. Its covariance is given by the inverse of a design matrix $\mathcal{I}_{i,a}(z_{a}^*)$, which is well-defined as long as $|A| \neq 1$ and the arm is observed at least twice.
The marginal estimator, $\hat{\theta}_{i,a}$, is therefore also an unbiased sub-Gaussian variable with an explicitly known variance. While the individual estimators are not strictly normal, their independence across users allows us to apply the Lindeberg-Feller Central Limit Theorem \citep{van2000asymptotic}. This demonstrates that the pooled estimator $\hat{\theta}_a$, obtained by averaging across a large number of users $N$, is asymptotically normal. This key result, combined with its reduced variance, provides the foundation for valid statistical inference.

Note that if we know the true value of $x_{i,a,0}$, then the variance of $\hat\theta_{i,a}$ will solely be $\frac{\sigma^2}{D^2|\mathcal{T}_{i,a}|}$, the value ${\left( \sum_{t \in \mathcal{T}_{i,a}} A^t \right)^2}/{\sum_{t \in \mathcal{T}_{i,a}} (A^t)^2}$ subtrancating from the dominator of variance of $\hat{\theta}_{i,a}$ in Theorem \ref{individual-MLE} is the infulence brought by jointly estimating the unknown intial state $x_{i,a,0}$. The influence of not knowing the true $x_{i,a,0}$ on estimating $\theta_a$ has an absolute upper bound illustrated by this proposition:

\begin{proposition}\label{fisherBound}
Let $\mathcal{T}$ be a finite set of positive integers, and let $A \in \mathbb{R}$ with $|A| \neq 1$ and $A\neq 0$. Then
$\frac{\left(\sum_{t \in \mathcal{T}} A^t\right)^2}{\sum_{t \in \mathcal{T}} A^{2t}} \leq \frac{1 + |A|}{|1 - |A||}.$
\end{proposition}

The proof proceeds by utilizing properties of geometric series to bound the ratio, showing that it is always at most $\frac{1 + |A|}{|1 - |A||}$. Proposition \ref{fisherBound} demonstrates that the increase in variance from not knowing the initial state is bounded by a constant that depends only on the system dynamics. The practical utility of this result is that it helps simplify variance calculations and analysis. Instead of requiring the specific set of observation times $\mathcal{T}_{i,a}$ to calculate the exact variance, we can use this bound to establish a reliable worst-case variance that depends only on the \textit{number} of samples, $|\mathcal{T}_{i,a}|$. This clarifies that the dominant factor for driving the estimator variance to zero is indeed the sample size. To ensure reliable estimation and sufficient statistical power for all arms, our primary goal must therefore be to guarantee that each $|\mathcal{T}_{i,a}|$ is sufficiently large.

This directly motivates our approach to policy design. To prevent any arm from being chosen too infrequently, we must enforce a minimum level of exploration. We formalize this by introducing a minimum exploration probability, $p_{\min}$, which serves as a lower bound on the probability of selecting any arm, for any user, at any point in time. By setting $p_{\min} > 0$, we ensure that, in expectation, the number of samples for each arm grows linearly with time.

\subsection{Deriving minimum exploration probability}

The importance of controlling variance becomes evident when considering power. For example, to detect a treatment effect difference between two arms $a$ and $a'$ with high probability, the variances of their respective estimators must be small. The following result makes this tradeoff explicit:

\begin{proposition}
\label{betaCalculation}
Consider a hypothesis test comparing two treatment effects, $\theta_a$ and $\theta_{a'}$, for actions $a, a' \in \mathcal{A}$, with the null hypothesis $H_0: \theta_a = \theta_{a'}$ and the alternative hypothesis $H_1: \theta_a > \theta_{a'}$. The Type II error $\beta$ for detecting a treatment effect difference of $\Delta = \theta_a - \theta_{a'}$ is given by:
$\beta = \Phi\left(\Phi^{-1}(1 - \alpha) - \frac{\Delta}{\sqrt{\var(\hat{\theta}_a) + \var(\hat{\theta}_{a'})}}\right),$
where $\alpha$ is the Type I error rate, $\Phi$ is the cumulative distribution function of the standard normal distribution, and $\var(\hat{\theta}_a)$ and $\var(\hat{\theta}_{a'})$ denote the variances of the estimators for $\theta_a$ and $\theta_{a'}$, respectively.
\end{proposition}

For a large number of users $N$, both $\hat{\theta}_a$ and $\hat{\theta}_{a'}$ are asymptotically normal. Their difference is also also approximately normal with variance $\var(\hat{\theta}_a) + \var(\hat{\theta}_{a'})$. This large-sample approximation allows for the construction of a test statistic based on the standardized difference. The Type II error $\beta$ follows from evaluating the probability that this statistic is below $\Phi^{-1}(1 - \alpha)$ under the alternative. The equation in Proposition \ref{betaCalculation} illustrates that, for fixed Type I error $\alpha$ and effect size $\Delta$, reducing the estimation variances directly increases the power of the test. Thus, any algorithm used to generate action probabilities must be designed to provide enough information for each arm to ensure sufficiently small estimator variance. Because multiple pairwise hypothesis tests are conducted when comparing each arm against a baseline, it is important to control the overall (family-wise) Type I and Type II error rates to maintain statistical validity across all comparisons. The family-wise error rate refers to the probability of making at least one false conclusion among the set of tests. Specifically, if there are $|\mathcal{A}| - 1$ pairwise tests for each arm against the baseline, controlling the family-wise error rates $\alpha_0$ and $\beta_0$ requires adjusting the individual test error rates accordingly. The following theorem formalizes these requirements by providing a lower bound on $p_{\min}$, along with an upper bound on the maximum selection probability $p_{\max}$, ensuring that each arm is sampled sufficiently to guarantee valid inference.

\begin{theorem}
\label{piMin}
Let $\Delta_0$ be the minimal detectable effect size. Suppose we aim to control the family-wise Type I error rate at level $\alpha_0$ and the family-wise Type II error rate at level $\beta_0$ when testing all pairwise treatment effect differences between each arm $a \in \mathcal{A} \setminus \{a'\}$ and the baseline arm $a'$.

By applying the Bonferroni correction \citep{holm1979simple}, each individual pairwise test requires a Type I error rate $\alpha$ and a Type II error rate $\beta$ such that
$\alpha_0 \ge (|\mathcal{A}| - 1) \alpha, \beta_0 \ge (|\mathcal{A}| - 1) \beta,$
then the minimum exploration probability $p_{\min}$ per arm must satisfy
     $   p_{\min} \geq \frac{1}{T} \left( \frac{1 + |A|}{|1 - |A||} + \frac{2\sigma^2 \left( \Phi^{-1}\left(1 - \frac{\alpha_0}{|\mathcal{A}| - 1} \right) - \Phi^{-1}\left( \frac{\beta_0}{|\mathcal{A}| - 1} \right) \right)^2}{ND^2\Delta_0^2} \right).$
Furthermore, to maintain a valid probability distribution over arms, the maximum selection probability of any single arm is bounded above by
     $   p_{\max}\leq 1-(|\mathcal{A}|-1)p_{\min}
        .$
\end{theorem}

We first apply Proposition~\ref{fisherBound} to upper bound the uncertainty in estimating $\theta_a$ due to the unknown initial state $x_{i,a,0}$. If each arm is selected with probability at least $p_{\min}$, then over $T$ rounds, the number of pulls for each arm provides a lower bound on information and hence an upper bound on the variance of each user-level estimator.
By Theorem~\ref{individual-MLE}, these individual variances combine to give an upper bound on the variance of the population-level estimator $\hat{\theta}_a$. This variance must fall below a threshold determined by the desired error rate and effect size in order to achieve sufficient power (Proposition~\ref{betaCalculation}).
Since there are $|\mathcal{A}| - 1$ pairwise tests, we apply Bonferroni correction to control the family-wise error rate. This introduces a constraint on $p_{\min}$, and the upper bound $p_{\max}$ follows from the constraint that the total assignment probability must sum to at most one: $p_{\max} \leq 1 - (|\mathcal{A}| - 1)p_{\min}$.

\subsection{Clipping algorithm and regret analysis}

With the lower bound on $p_{\min}$ and the corresponding upper bound on $p_{\max}$ established, we now turn to the practical question of how to enforce these constraints within bandit algorithms. Let $p_{i,a,t}$ denote the probability that user $i$ is assigned to arm $a \in \mathcal{A}$ at time $t$ under the original (unconstrained) selection strategy $\Pi$.  
Common algorithms such as UCB and TS generate these probabilities based on current reward estimates. Specifically, UCB deterministically selects the arm with the highest confidence bound, assigning it probability 1 and all others 0. TS, on the other hand, determines the probability of each arm by the chance that a sample drawn from that arm’s posterior distribution is larger than samples from all other arms. However, neither method guarantees that the resulting probabilities satisfy the minimum and maximum probability constraints.

To ensure sufficient exploration while keeping the modified distribution close to the original, we define a clipped distribution $p_{i,a,t}'$ under the constrained policy $\Pi'$ developed from $\Pi$, by solving the following optimization problem. This projects the original probability vector onto the set of valid distributions bounded by $[p_{\min}, p_{\max}]$.

\begin{definition}[Clipped Probability Distribution]\label{clipProcess}
\textit{For each user $i$ at time $t$, the clipped probability distribution $p_{i,a,t}'$ is defined as the solution to the following optimization problem, which enforces each probability to lie within $[p_{\min}, p_{\max}]$ while remaining as close as possible to the original distribution $p_{i,a,t}$:}
\begin{equation}
\min_{p'} \big\{ \sum_{a \in \mathcal{A}} | p_{i,a,t} - p_{i,a,t}' | : \sum_{a \in \mathcal{A}} p_{i,a,t}' = 1, p_{\min} \leq p_{i,a,t}' \leq p_{\max} \  \forall a \in \mathcal{A} \big\}
\label{eq:clip_opt}
\end{equation}
\textit{This optimization problem is a linear program and admits an optimal solution as long as the feasible set is nonempty, which holds when $p_{\min} \leq {1}/{|\mathcal{A}|} \leq p_{\max}$.}
\end{definition}
To implement the clipping of probabilities in practice, we propose the following procedure. At each round and for each user, we first compute the base action probabilities from the chosen bandit algorithm, then apply the clipping optimization to enforce the minimum and maximum probability constraints before sampling the action. The complete steps are summarized in Algorithm \ref{ClippAlgo}.

The probability clipping algorithm is designed to enforce the desired constraints on action selection probabilities. While this procedure modifies the original policy, it is important to understand its impact on cumulative regret. The following result generalizes Theorem 4 from \cite{yao2021power}, extending it from the two-arm case to our setting with multiple arms. Recall that $R_{\Pi}(T)$ denotes the cumulative regret of the original (unclipped) policy $\Pi$ over a time horizon $T$ (per user), and let $R_{\Pi'}(T)$ denote the cumulative regret under the clipped policy $\Pi'$, where action probabilities are constrained to lie within $[p_{\min}, p_{\max}]$.

\SingleSpacedXII
\begin{algorithm}[H]
\caption{Probability Clipping Algorithm}
\label{ClippAlgo}
\begin{algorithmic}
\Require Action set $\mathcal{A}$, number of users $N$, constraints $p_{\min}$, $p_{\max}$, time horizon $T$
\State Initialize model parameters for each user $i \in \{1,\dots,N\}$ (e.g., prior for TS)
\For{$t = 1$ to $T$}
    \For{$i = 1$ to $N$}
        \State \textbf{Compute probabilities:}
        \State Compute base probabilities $p_{i,a,t}$ for all $a \in \mathcal{A}$ using the selected algorithm
        \State \textbf{Clip probabilities:}
        \State Solve the optimization in Eq.~\eqref{eq:clip_opt} to obtain clipped probabilities $p_{i,a,t}'$
        \State Sample an arm $a_{i,t} \sim p_{i,\cdot,t}'$ from categorical distribution and observe reward $r_{i,a_{i,t},t}$
        \State Pass $r_{i,a_{i,t},t}$ and $a_{i,t}$ to the original algorithm, and update user $i$'s model parameters
    \EndFor
\EndFor
\end{algorithmic}
\end{algorithm}
\OneAndAHalfSpacedXII
\begin{theorem}\label{clipReg}
Suppose the original policy $\Pi$ achieves a regret rate $R_{\Pi}(T)$. Then, under the clipping procedure, the regret of the resulting clipped policy $\Pi'$ satisfies $R_{\Pi'}(T) \leq R_{\Pi}(p_{\max} T) + (1 - p_{\max}) C_gT.$
\end{theorem}

The proof follows the approach of \citet{yao2021power}. The clipping procedure modifies the original policy’s decisions on at most $(1 - p_{\max}) T$ steps, while on the remaining $p_{\max} T$ steps, the clipped policy $\Pi'$ agrees with $\Pi$. Since the per-step regret is bounded by $C_g$, the regret accumulated during clipped steps is at most $(1 - p_{\max}) C_g T$. On the other hand, the regret on the steps where the policies agree is at most $R_{\Pi}(p_{\max} T)$. This result implies that as $p_{\min}$ decreases with increasing $N$, the effect of clipping becomes negligible, and the policy asymptotically achieves improved or better regret performance.
We now apply this general regret bound to a specific instance: the clipped ROGUE-TS algorithm, which incorporates the clipping procedure into the ROGUE-TS framework to enforce the desired probability constraints.

\begin{corollary}
    The regret of the clipped ROGUE-TS algorithm satisfies the bound:
 $ R_{\Pi}'(t)\leq  \frac{8}{3}L_g^2 |\mathcal{A}|^{\frac{1}{4}}(p_{\max}T)^{\frac{3}{4}}\sqrt{ 2\sigma c_f(d_x,d_\theta) + 4L_p\sigma^2\sqrt{\log(p_{\max}T)}}
 +2C_g\min\{T,|\mathcal{A}|\}  + (1 - p_{\max}) C_gT.$
\end{corollary}

This result highlights several desirable properties of the clipped ROGUE-TS policy. First, it retains a near-optimal regret bound, deviating from the original only by an additive linear term. Moreover, this term vanishes as we add additional study participants or extend the study timeline. Second, the clipping mechanism introduces controlled exploration by enforcing minimum and maximum selection probabilities. Finally, since the algorithm remains randomized after clipping, it naturally aligns with the requirements of MRT-style inference, where probabilistic treatment assignment is essential for valid inference.

\section{Numerical experiments}
In this section we consider two numerical studies that validate the performance of our approaches using both real-world and synthetic datasets. The first set of experiments examines the regret properties of ROGUE-TS compared to state-of-the-art non-stationary bandit methods. Here we use an MRT dataset on workplace sedentary behavior, with a complementary bipolar disorder MRT dataset reported in Appendix~\ref{bdSimulation}. The second set of experiments focuses on the capabilities of our probability clipping approach in achieving sufficient statistical power, again leveraging the workplace sedentary MRT dataset.

\subsection{Numerical experiments for ROGUE-TS}\label{sec:ExperimentsTS}
In this section, we present three numerical experiments to evaluate the performance of the proposed ROGUE-TS algorithm against several established nonstationary bandit baselines. These experiments are designed to assess both long-run reward and the speed of adaptation to nonstationary environments. All experiments were implemented in Python 3.10 and run on a laptop computer with a 2.4GHz processor and 16GB RAM. The first experiment uses synthetically generated data under a generalized linear model (GLM) to evaluate algorithmic performance in a controlled setting. The second and third experiments apply ROGUE-TS to real-world mobile health contexts: an MRT focused on improving workplace activity levels through behavioral messaging, and a bipolar disorder MRT experiment reported in Appendix~\ref{bdSimulation}.

\subsubsection{Experiment 1: Simulation with Gaussian ROGUE-GLM}\label{ToyReg}
Our first experiment is based on a synthetic dynamic GLM model \citep{mccullagh2019generalized}, which can be used to extend classical stationary MAB frameworks to account for time-varying rewards \citep{filippi2010parametric}. The exact dynamics used in the simulation for each $a\in \mathcal{A}$ are as follows: the transition function is given by
$h_a(x_t, \pi_t) = A_{a} x_t + B_{a} \pi_t + K_{a},$
where $A_a$, $B_a$, and $K_a$ matrices/vectors of appropriate dimensions, and the rewards $r_{a,t}$ follow a Gaussian distribution with expectation governed by a generalized linear model:
$\mathbb{E}[r_{a,t}] = g_a(\theta_a, x_{a,t}) = \alpha_a^\top \theta_a + \beta_a^\top x_{a,t}$, and variance $\sigma_a^2$.
While the theoretical framework permits general parameter spaces, we simplify the simulation by treating $x$ and $\theta$ as real-valued scalars (i.e., $x, \theta \in \mathbb{R}$) and all related parameters, including vectors and matrices such as $A_a$, $B_a$, and $K_a$ as scalars. For illustrative purposes, all scalar parameters, including $x_0$, $\theta_a$, $A_a$, $B_a$, $K_a$, $\alpha_a$, $\beta_a$, are independently drawn from the uniform distribution on $[0,1]$, while $\sigma_a$ is drawn from $[2,3]$. Consequently, the sets $\mathcal{X}$ and $\Theta$ are instantiated as $[0,1]$ for the simulation. Since these values are randomly generated and not fixed across experiments, we omit a parameter table. To enable sequential Bayesian updates in this linear-Gaussian setting, we apply a Kalman filter per arm over the joint state $ (\theta_a, x_{a,t})$. For compatibility with the Kalman framework, we reformulate the system accordingly; the detailed procedure is provided in Appendix~\ref{app:kalman-rogue}.

\begin{figure}[H]
    \centering
    \begin{minipage}{0.35\textwidth}
        \centering
        \includegraphics[width=\textwidth]{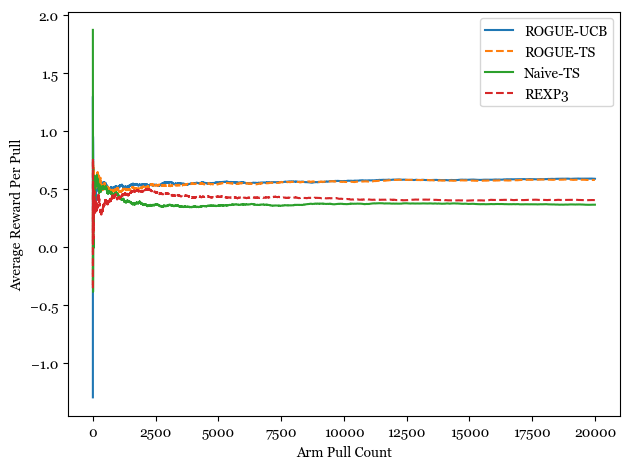}
        \caption{Average reward for Gaussian ROGUE simulation}
    \label{fig:reward of GRS}
    \end{minipage}
    \begin{minipage}{0.35\textwidth}
        \centering
        \includegraphics[width=\textwidth]{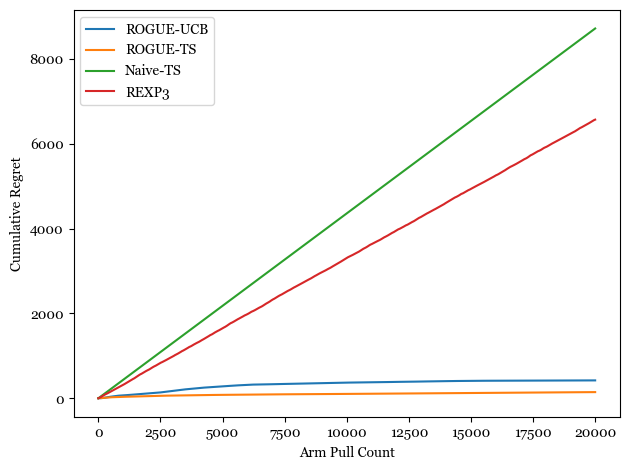}
        \caption{Cumulative regret for Gaussian ROGUE simulation}
    \label{fig:regret of GRS}
    \end{minipage}
\end{figure}

We compare ROGUE-TS with three baselines: the first is naive Thompson Sampling \citep{Thompson1933}, which does not account for nonstationarity; the second is REXP3 that makes decisions by assigning a probability distribution to each action, which is updated based on the observed rewards utilizing exponential weighting to prioritize actions with higher cumulative rewards over time \citep{rexp3}; the third baseline is the ROGUE-UCB algorithm which has been shown be effectively balance exploration and exploitation in nonstationary environments \citep{mintz2020nonstationary}. Each algorithm was run for 20,000 rounds and repeated across 20 trials. Figures~\ref{fig:reward of GRS} and~\ref{fig:regret of GRS} display the average reward per arm pull and cumulative regret, averaged over trials. Moreover, hyper parameters for each algorithm where set to their theoretical optimal amounts.

As shown in the figures, both ROGUE-TS and ROGUE-UCB consistently outperform the baselines. ROGUE-TS achieves the lowest regret and highest reward, demonstrating its ability to adapt quickly and efficiently to nonstationary dynamics. ROGUE-UCB also performs well, while REXP3 lags behind due to its conservative adaptation strategy. Naive-TS, which does not model nonstationarity, performs the worst, highlighting the importance of incorporating dynamic structure into the learning algorithm.

\subsubsection{Experiment 2: Micro-randomized trial for workplace activity}\label{MRTregExpr}

The second experiment is based on a MRT conducted in a workplace setting, aiming to reduce prolonged sitting through timely behavioral interventions \citep{Carpenter2025TransitionISBNPA,CarpenterInPrepTransitionPilot}. Participants received contextually delivered messages every 30 minutes during an 8-hour workday, encouraging them to either stand or walk. The action space consists of three message types: (1) no message, (2) a prompt to stand up, and (3) a prompt to move (e.g., take a short walk). Rewards are defined as the decrease in sitting time during the 30 minutes following each message compared to the 30 minutes prior, capturing the immediate behavioral impact of each intervention. The decision environment is nonstationary, reflecting daily variation in employee receptivity and behavioral fatigue or habituation effects over time. We applied Expectation Maximization (EM) \citep{rabiner2002tutorial} to infer system parameters from longitudinal data, using the model described in Section~\ref{ToyReg} with three arms. EM was applied to historical message and sitting-time data to learn the parameters of a linear Gaussian system, enabling the simulation of realistic user behavior under varying intervention strategies. Each trial simulates 1,000 arm pulls, corresponding to approximately 1 month of intervention assuming a standard 5-day workweek and one decision every 30 minutes over 8-hour workdays. As in prior experiments, we compare the performance of ROGUE-TS with REXP3, naive TS, and ROGUE-UCB, evaluating both long-run reward and regret over 20 replicated trials.

\begin{figure}[H]
    \centering
    \begin{minipage}{0.35\textwidth}
        \centering
        \includegraphics[width=\textwidth]{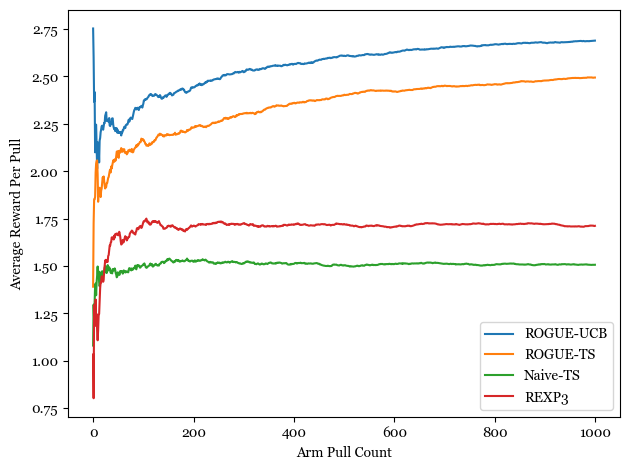}
        \caption{Average reward for MRTs simulation}
        \label{fig:reward mrt}
    \end{minipage}
    \begin{minipage}{0.35\textwidth}
        \centering
        \includegraphics[width=\textwidth]{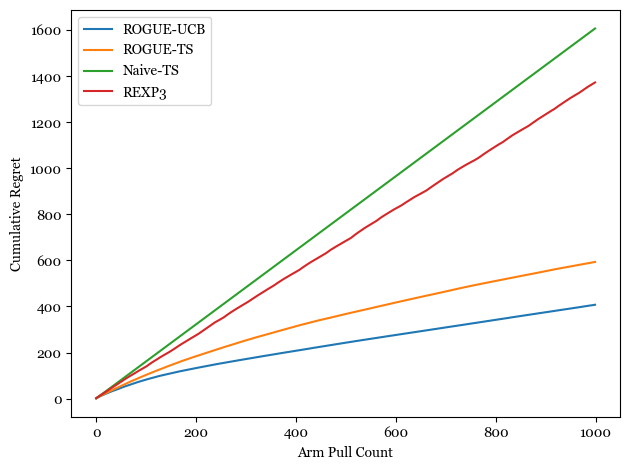}
        \caption{Cumulative regret for MRTs simulation}
        \label{fig:regret mrt}
    \end{minipage}
\end{figure}

As shown in Figures~\ref{fig:reward mrt} and~\ref{fig:regret mrt}, ROGUE-TS consistently achieves higher rewards and lower regret than baseline methods, demonstrating effective adaptation to user behavior in this dynamic setting. Notably, ROGUE-TS outperforms the next best algorithm by nearly one additional minute of standing time within a 30-minute interval. Although this increase may appear modest, prior research has linked even small increments in standing time to meaningful health benefits \citep{pereira2020efficacy,toledo2023does}. For example, replacing just 30 minutes of sitting per day with standing can improve metabolic markers such as glucose and insulin levels, reduce cardiovascular risk factors, and lower all-cause mortality risk \citep{owen2010too}. Therefore, an extra minute of standing per half-hour intervention could cumulatively yield significant long-term health improvements, particularly in sedentary workplace populations. While ROGUE-TS demonstrates strong overall performance, it converges more gradually than ROGUE-UCB, which explains why their reward remains close throughout the trials. However, by the end of the 1000-trial horizon, the regret of ROGUE-TS approaches that of ROGUE-UCB, indicating that ROGUE-TS is closing the gap in long-term learning efficiency.

\subsection{Numerical experiments for power constraints}
Following the verification of the effectiveness of the ROGUE-TS algorithm, this section presents two experiments evaluating its power-constrained variant, focusing on its ability to detect meaningful intervention effects while maintaining strong decision-making performance. The first experiment revisits the synthetic GLM-based environment in Section \ref{ToyReg}, now designed according to the model specified in Section \ref{PCModel}, with predefined intervention effects. We investigate the trade-off between statistical inference performance and individual-level reward optimization. 
Specifically, we measure empirical Type I and Type II error rates, as well as the regret ratio between the original and clipped algorithms. The second experiment builds on the MRT simulations described in Experiment \ref{MRTregExpr}. We reformulate the environment using the model in Section \ref{PCModel} and re-run the EM algorithm to estimate a shared treatment effect while accounting for participant-specific dynamics. We then conduct simulations using this reformulated model and evaluate both statistical power and the regret ratio, providing insights into the algorithm’s performance in a realistic mobile health context.

In this set of experiments, we change the policies under comparison from previous sections and focus on evaluating the performance gain of our algorithms under power constraints relative to a \textit{Full Exploration} baseline, which selects arms uniformly at random at each time step. 
Specifically, we include the ROGUE-TS and ROGUE-UCB algorithms along with their clipped variants.

\subsubsection{Power testing on Gaussian ROGUE-GLM}\label{PowerToy}

Using the model specified in Section \ref{PCModel}, we conducted a synthetic simulation involving 15 participants and a 3-arm design. For each participant $i \in \{1, 2, \dots, 15\}$, the state dynamics follow the transition function $h_{i,a}(x_{i,t}, \pi_{i,t}) = A_{i,a} x_{i,t} + B_{i,a} \pi_{i,t} + K_{i,a}$, where $A_{i,a}$, $B_{i,a}$, and $K_{i,a}$ are scalars. The rewards $r_{i,a,t}$ are Gaussian with expected values defined by a link function of the form $ \mathbb{E}[r_{i,a,t}] = g_{i,a}(\theta_{a}, x_{i,a,t}) = \theta_a + \beta_{i,a} x_{i,a,t}$ and variance $\sigma_{i,a}^2$. The sets $\mathcal{X}$ was defined as $[0,1]$ and $\Theta$ was defined as $[0,1]$. This setup is similar to the experiment in Section~\ref{ToyReg}, where the dynamics were randomly generated. However, in the current experiment, for illustrative purposes and to ensure identifiability, we generate $A_{i,a}$ within the range $[0, 0.9]$ and $\sigma_{i,a}$ within the range $[1,1.5]$. Additionally, the parameter $\theta$ is predefined and fixed across participants, with $\theta=0.5$ for arms 1 and 3 and $\theta=1$ for arm 2. We interpret arm 1 as \textit{do nothing}, while arms 2 and 3 correspond to distinct active interventions.

Using Theorem~\ref{piMin}, we compute $p_{\min}$ under a family-wise Type I error rate of $\alpha_0 = 0.05$, a Type II error rate corresponding to a desired family-wise power of $1 - \beta_0 = 0.8$, and a target effect size of $\Delta_0 = 1$, with a time horizon of $T = 90$. We allow the parameters $A$ and $\sigma$ to vary across arms and users. Therefore, $p_{\min}$ is computed for each possible $(A, \sigma)$ pair using Theorem~\ref{piMin}, and for each user, we select the maximum $p_{\min}$ value across all arms to ensure sufficient exploration under the most demanding conditions. 

\begin{figure}[H]
    \centering
    \begin{minipage}{0.35\textwidth}
        \centering
        \includegraphics[width=\textwidth]{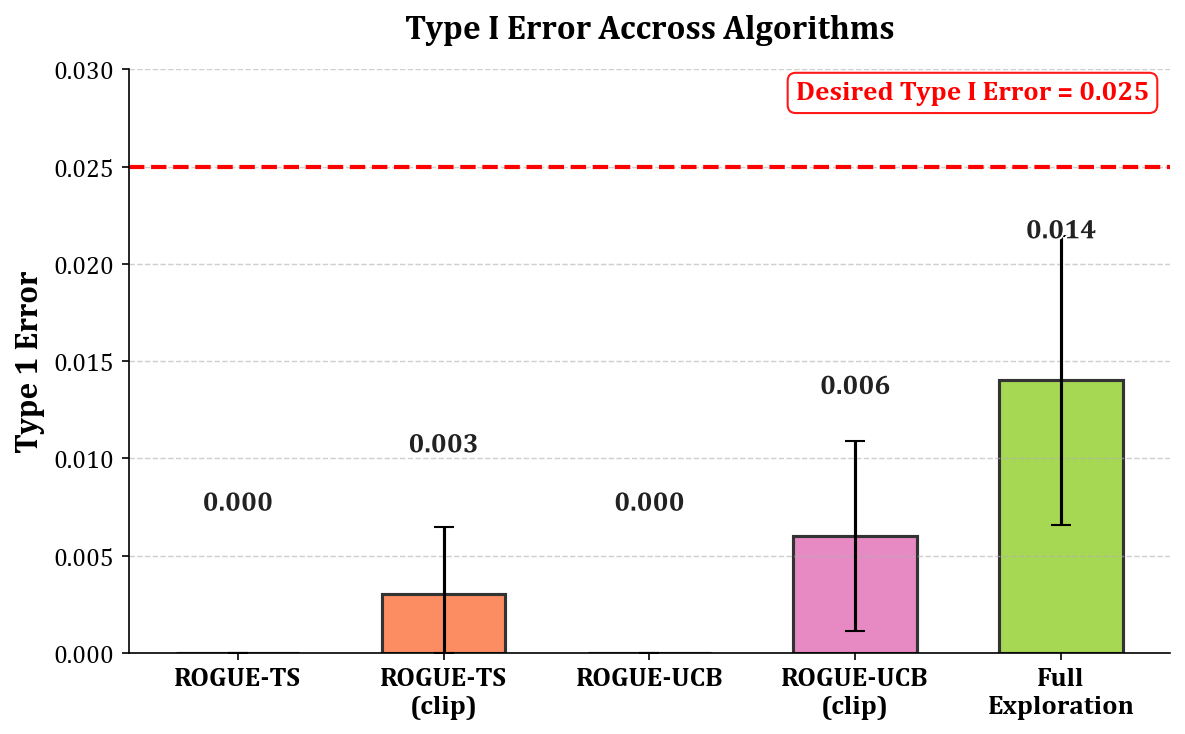}
        \caption{Type I Error on ROGUE-GLM}
        \label{fig:type1Toy}
    \end{minipage}
    \
    \begin{minipage}{0.35\textwidth}
        \centering
        \includegraphics[width=\textwidth]{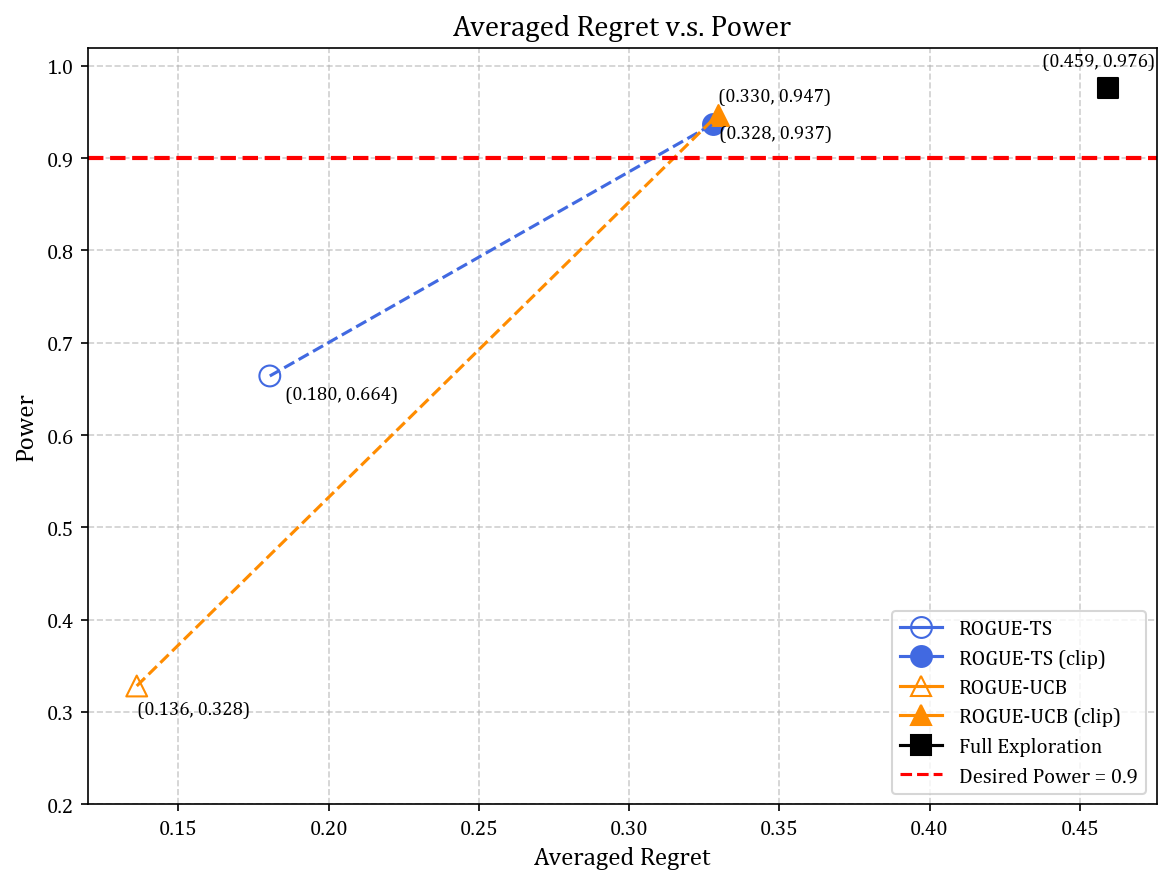}
        \caption{Average Regret v.s. Power on ROGUE-GLM}
        \label{fig:powerToy}
    \end{minipage}
\end{figure}

Since Actions 1 and 3 were predefined to have identical effects, we examine the Type I error rate by comparing outcomes between these two arms. Due to multiple testing over the action set $\mathcal{A}$ with $|\mathcal{A}| = 3$, we apply a Bonferroni correction, adjusting the significance level to $\alpha = {0.05}/({|\mathcal{A}| - 1} )= 0.025.$
Action 2 has a significantly higher effect than Actions 1 and 3, so we evaluate the Type II error rate by comparing Actions 1 and 2, with adjusted target
$\beta = {0.2}/({|\mathcal{A}| - 1}) = 0.1.$ According to the Bonferroni correction \citep{holm1979simple}, if the empirical error rates fall below these adjusted thresholds, the power constraints are satisfied.
We assess the cumulative regret ratio between clipped and original algorithms, averaged over all 10 participants and 1,000 replications. For each replication, we record whether Type I or Type II errors occur and compute empirical error rates. The results of this experiment are presented in Figures~\ref{fig:type1Toy} and \ref{fig:powerToy}, showing the Type I error rates and the relationship between average regret and statistical power for ROGUE-GLM.

In terms of the tradeoff between statistical power and average regret per pull, the clipped ROGUE policies consistently achieve power above 0.9 while maintaining substantially lower average regret per pull, around 0.95, compared to the Full Exploration baseline. Although the Full Exploration strategy yields the highest power, it also incurs the highest average regret per pull. In contrast, the original unclipped ROGUE policies achieve lower average regret per pull but fail to meet the desired power threshold of 0.9. These results demonstrate that clipping significantly improves statistical power while introducing only a moderate increase in average regret per pull, which is consistent with the regret scaling behavior predicted by Theorem~\ref{clipReg}.

Interestingly, ROGUE-UCB (clip), ROGUE-TS (clip) and Full Exploration showed nonzero empirical Type I error, while ROGUE-TS and ROGUE-UCB had none in our simulations. This indicates that the higher power of the former methods may come with slightly relaxed Type I error control. Still, all observed error rates remain acceptable. Importantly, zero Type I error does not imply perfect control, but rather a bound limited by simulation size. Overall, clipped ROGUE policies strike a strong balance between power, regret, and error control.

\subsubsection{Power Testing on workplace MRT}

Using the model specified in Section \ref{PCModel}, we re-ran the EM algorithm to estimate the shared treatment effects and individual dynamics. In this analysis, we set the range of $\theta$ to $[0,2]$. The estimated shared effects are $\theta=0.3936$ for \textit{No Message}, $\theta=1.1458$ for \textit{Stand Prompt}, and $\theta=0.8449$ for \textit{Move Prompt}.

This experiment follows the same dynamic setup as in Experiment~\ref{PowerToy}, using parameters estimated via the EM algorithm. We apply Theorem~\ref{piMin} to determine the required minimum intervention probability, targeting a family-wise Type I error rate of $\alpha_0 = 0.05$ and a family-wise power of $1 - \beta_0 = 0.8$ to detect an effect size of $\Delta_0 = 0.4$. With a time horizon of $T = 210$ (where the larger value of $T$ is motivated by the relatively high variance), $p_{\min}$ is computed for each user-action pair based on their estimated $(A, \sigma)$ values. For each user, we then select the maximum $p_{\min}$ across all arms to ensure adequate exploration under the most challenging condition. We then apply a Bonferroni correction, setting the per-test significance level to $\alpha = 0.025$ and the Type II error rate to $\beta = 0.1$, consistent with the approach used in Experiment~\ref{PowerToy}. Since the treatment effects for both interventions exceed the meaningful effect size of 0.4 relative to the `No Message' condition, our focus is on evaluating the family-wise power of different algorithms.

\begin{figure}[H]
    \centering
    \includegraphics[width=0.3\textwidth]{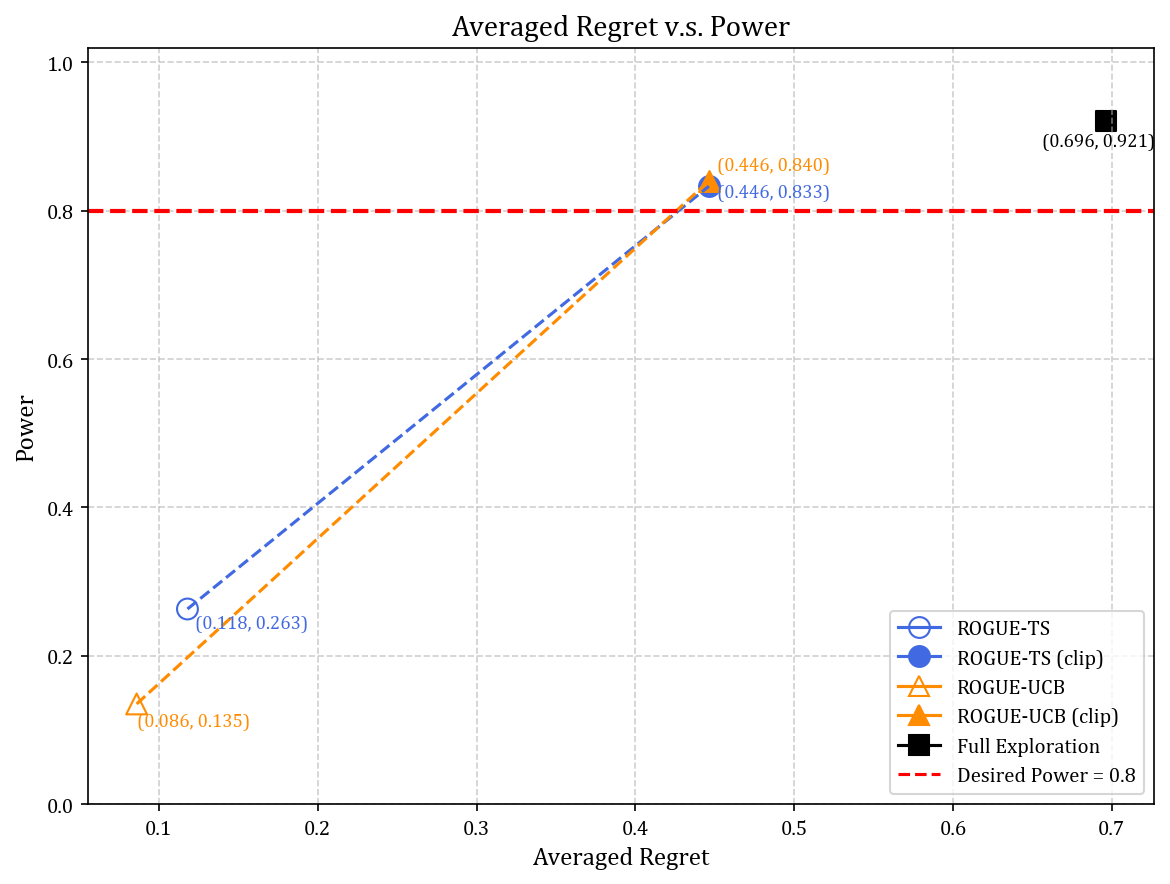}
        \caption{Family-Wise Joint Power vs. Averaged Regret}
        \label{fig:type2mrt_compare}
\end{figure}

Similar to the findings in Experiment~\ref{PowerToy}, our MRT experiment shows that unclipped algorithms achieve low regret but fail to meet the statistical power constraint, limiting their usefulness for intervention effect detection. In contrast, the clipped ROGUE variants maintain low regret while satisfying the power requirement and outperforming the Full Exploration baseline in terms of individual-level performance. Beyond numerical gains, clipped ROGUE algorithms offer practical advantages in real-world MRTs by enabling valid inference without compromising personalization. Unlike uniform exploration, which often applies suboptimal actions, our context-aware, structured exploration adapts to nonstationary user behavior while ensuring reliable effect estimation. This principled balance between learning and optimization makes clipped ROGUE a strong candidate for effective MRT design.

\section{Managerial implications}\label{sec:Results}

For researchers and practitioners conducting MRTs for digital health interventions, our modeling framework and theoretical findings offer practical guidance to improve trial design and execution. First, our results support that a Bayesian approach such as ROGUE-TS can effectively leverage prior data, such as observational records or pilot study outcomes, to inform the initialization of adaptive treatment policies. Rather than relying solely on uniform randomization, which may waste decision points on ineffective options, researchers can use historical engagement patterns or baseline user behavior to optimize a JITAI. By doing so, MRT studies can be designed that minimize burden and meet the needs of individual participants.

Second, when interventions vary in burden or risk, maintaining a high level of exploration during treatment assignment can lead to suboptimal short-term outcomes. For example, in physical activity promotion, low-cost prompts like standing reminders are less disruptive than longer walking suggestions, which may interfere with work. In digital mental health applications, certain messages might unintentionally trigger adverse reactions. In such contexts, carefully balancing exploration and exploitation is crucial. Since our results are conservative, implying that the derived $p_{\min}$ is higher than strictly necessary, practitioners may consider lowering the exploration ratio further in higher-risk settings to prioritize safety. Conversely, when the risk is low, using our $p_{\min}$ ensures strong inference performance while still maintaining adequate protection against poor estimation.

Third, our approach allows researchers to efficiently tradeoff the number of participants ($N$) and the duration of the trial ($T$) to achieve reliable inference. As shown by our analysis, increasing either sample size or follow-up time improves statistical power but entails the need for greater recruitment and management costs. Analytically, this is because $N$ and $T$ directly determine $p_{\min}$. When study managers can recruit additional participants, exploration can be reduced, thereby increasing personalization and lowering the level of randomization imposed on individuals. When recruitment is constrained, extending the trial horizon offers another way to maintain significant conclusions. In practice, this means that our approach provides managers with clear levers to manage costs, minimize disruption, and still preserve the validity of personalization strategies.

\section{Extensions and limitations}

Despite the promising results, our work has several limitations that motivate future research.

\begin{enumerate}

    \item For more complex dynamics or reward distributions, exact Bayesian posterior updates become more challenging. Extensions based on approximation methods, such as particle filters, and analysis of their impact on regret represent promising directions for future research. However, results like Proposition~\ref{decom}, which rely on exact posterior updates, may no longer hold, thereby weakening the our current regret bounds. A promising approach is to develop regret bounds based on divergence measures between approximate and exact posteriors, as proposed in \cite{phan2019thompson}.

    \item While $p_{\min}$ ensures a lower bound on exploration for valid inference, a fixed threshold can be conservative, particularly after sufficient evidence has been accumulated, or when fast-changing dynamics naturally induce action variation even under exploitation. A promising direction is to design adaptive clipping strategies that relax this constraint based on inference-aware criteria. For example, one could extend the algorithm to monitor the variance of estimators $\hat{\theta}_{i,a}$ or $\hat{\theta}_a$, and reduce $p_{\min}$ as the variance falls below a threshold derived from power analysis.

    \item Although fairness is not a primary concern in our current context, where the focus is on optimization and inference for health intervention settings, it becomes more important in broader deployment scenarios. Adaptive algorithms in the ROGUE style may inadvertently disadvantage certain arms, especially in settings where ensuring fairness across actions or individuals is critical. One promising direction is to maintain confidence intervals over the states and dynamically adjust exploration based on whether the induced reward intervals overlap, following a similar idea to that proposed in \cite{joseph2016fairness}.
\end{enumerate}

\section{Conclusion}\label{sec:Conclusion}
In this paper, we develop a novel Thompson Sampling algorithm tailored for ROGUE bandits, termed ROGUE-TS, and establish theoretical guarantees demonstrating its sublinear regret growth. Empirical validation on two separate real-world datasets from MRTs, designed to optimize JITAIs aiming to prevent bipolar disorder and reduce workplace sedentary behavior, respectively, highlights the effectiveness of ROGUE-TS. These results demonstrate meaningful behavioral impacts and showcase the potential of our approach in mental health treatment, mobile health applications, and lifestyle behavior change interventions. To meet the dual goals of adaptive treatment allocation and rigorous statistical inference, especially in contexts like Micro-Randomized Trials, we propose a power-constrained extension of the ROGUE framework. This extension enforces a minimum intervention probability through probability clipping, balancing regret minimization with the statistical power necessary for valid causal inference. Our theoretical and empirical results confirm that this approach maintains high statistical power without substantially increasing regret, enabling reliable detection of treatment effects while accounting for complex individual dynamics.

\bibliographystyle{informs2014} 
\bibliography{ref} 

%
%
%
\newpage
\ECSwitch
\ECHead{E-Companion and Appendices}
\begin{APPENDICES}
\section{Proofs of Propositions in Text}
\phantomsection
\label{ap:proof}

\proof{Proof of Proposition~\ref{decom}}
The structure of this proof closely follows the argument in \cite{l2o}, adapted to our setting. 
    Consider a UCB algorithm with an upper confidence bound sequence $U = \{U_t|t \in N\}$. Note that, conditioned on $H_t$, the optimal action $\pi^*_t$ and the action $\pi_t$ selected by TS are identically distributed, and $U_t$ is deterministic. Hence, 
    $
    \mathbb{E}[U_t(\pi^*_t) \mid H_t] = \mathbb{E}[U_t(\pi_t) \mid H_t].
    $
    leading to:
\begin{align}
\mathbb{E}(r_{\pi^*_t,t}-r_{\pi_t,t}) &= \mathbb{E}\left(\mathbb{E}(r_{\pi^*_t,t}-U_t(\pi_t)+U_t(\pi_t)-r_{\pi_t,t} \mid H_t)\right)\\
&= \mathbb{E}\left(\mathbb{E}(r_{\pi^*_t,t}-U_t(\pi^*_t)+U_t(\pi_t)-r_{\pi_t,t} \mid H_t)\right)\\
&= \mathbb{E} (r_{\pi^*_t,t}-U_t(\pi^*_t))+\mathbb{E}(U_t(\pi_t)-r_{\pi_t,t}).
\end{align}
\endproof

\begin{remark}
    In our setting, this decomposition remains valid because the ROGUE-TS algorithm performs exact Bayesian inference at each step. Consequently, the selected action is Bayesian-optimal under the current posterior belief. This aligns with the framework in \cite{l2o}, which interprets TS as a stochastic maximizer of UCB-type indices. Thus, UCB-style upper bounds can be leveraged to analyze TS regret, even in nonstationary environments.
\end{remark}

\proof{Proof of proposition \ref{past-reward-kl-divergence}}

By Assumption~\ref{rewardAssum}, for each $ t' \in \mathcal{T}_{a,t} $, the reward is sub-Gaussian with parameter $ \sigma $, and the expectation of different parameter are given by 
$g(\theta_a, h_a^{t'}(x_{a,0}))$ and $g(\theta_a', h_a^{t'}(x_{a,0}')).$
Since the Gaussian distribution maximizes entropy under a fixed variance constraint, it minimizes the KL divergence among sub-Gaussian distributions with the same expectation gap \citep{bickel2015mathematical}. Therefore, we have the pointwise lower bound:
\begin{equation}
D_{\mathrm{KL}}(\mathbb{P}_{\theta_a,x_{a,0}}^{t'} \,\|\, \mathbb{P}_{\theta_a',x_{a,0}'}^{t'}) 
\geq \frac{1}{2\sigma} \left( g(\theta_a, h_a^{t'}(x_{a,0})) - g(\theta_a', h_a^{t'}(x_{a,0}')) \right)^2.
\end{equation}

Averaging over $ t' \in \mathcal{T}_{a,t} $, the trajectory KL divergence satisfies
\begin{equation}
\frac{1}{n(\mathcal{T}_{a,t})} D_{a,\pi_1^t}(\theta_a, x_{a,0} \,\|\, \theta_a', x_{a,0}')
\geq \frac{1}{2\sigma n(\mathcal{T}_{a,t})} \sum_{t'\in \mathcal{T}_{a,t}} \left( g(\theta_a, h_a^{t'}(x_{a,0})) - g(\theta_a', h_a^{t'}(x_{a,0}')) \right)^2. 
\end{equation}

\endproof

\proof{Proof of Lemma \ref{lemma1}}
Assume, for the sake of contradiction, that the inequality does not hold. Then there exist $\epsilon > 0$, $\theta \in \Theta$, and $x_0 \in \mathcal{X}$ such that
\begin{equation}
|g(\theta, h_a^t(x_0)) - g(\hat{\theta}_a, h_a^t(\hat{x}_{a,0}))| = L_g^2 \sqrt{2\sigma\gamma} + \epsilon.
\end{equation}

Since $g$ is bi-Lipschitz continuous with constant $L_g$, we have
\begin{equation}
\frac{1}{L_g} \|(\theta, h_a^t(x_0)) - (\hat{\theta}_a, h_a^t(\hat{x}_{a,0}))\|_2 \leq |g(\theta, h_a^t(x_0)) - g(\hat{\theta}_a, h_a^t(\hat{x}_{a,0}))| \leq L_g \|(\theta, h_a^t(x_0)) - (\hat{\theta}_a, h_a^t(\hat{x}_{a,0}))\|_2.
\end{equation}

Moreover, since $h$ is $L_h$-Lipschitz continuous with $L_h \leq 1$, and the Euclidean norm $\|\cdot\|_2$ is 1-Lipschitz, it follows that
\begin{equation}
\|(\theta, h_a^t(x_0)) - (\hat{\theta}_a, h_a^t(\hat{x}_{a,0}))\|_2 \leq \|(\theta, h_a^{t-1}(x_0)) - (\hat{\theta}_a, h_a^{t-1}(\hat{x}_{a,0}))\|_2.
\end{equation}

By recursively applying this for all $t' \in \mathcal{T}_{a,t}$ with $t' < t$, we get
\begin{equation}
\|(\theta, h_a^t(x_0)) - (\hat{\theta}_a, h_a^t(\hat{x}_{a,0}))\|_2 \leq \|(\theta, h_a^{t'}(x_0)) - (\hat{\theta}_a, h_a^{t'}(\hat{x}_{a,0}))\|_2.
\end{equation}

Combining these inequalities, for all $t' \in \mathcal{T}_{a,t}$, we have
\begin{align}
|g(\theta, h_a^{t'}(x_0)) - g(\hat{\theta}_a, h_a^{t'}(\hat{x}_{a,0}))| 
&\geq \frac{1}{L_g} \|(\theta, h_a^{t'}(x_0)) - (\hat{\theta}_a, h_a^{t'}(\hat{x}_{a,0}))\|_2 \\
&\geq \frac{1}{L_g} \|(\theta, h_a^t(x_0)) - (\hat{\theta}_a, h_a^t(\hat{x}_{a,0}))\|_2 \\
&\geq \frac{1}{L_g^2} |g(\theta, h_a^t(x_0)) - g(\hat{\theta}_a, h_a^t(\hat{x}_{a,0}))| \\
&= \frac{1}{L_g^2} \bigl(L_g^2 \sqrt{2\sigma\gamma} + \epsilon\bigr) = \sqrt{2\sigma\gamma} + \frac{\epsilon}{L_g^2}.
\end{align}

Hence, for every $t' \in \mathcal{T}_{a,t}$,
\begin{equation}
|g(\theta, h_a^{t'}(x_0)) - g(\hat{\theta}_a, h_a^{t'}(\hat{x}_{a,0}))| > \sqrt{2\sigma\gamma},
\end{equation}
which implies
\begin{equation}
\frac{1}{2\sigma n(\mathcal{T}_{a,t})} \sum_{t' \in \mathcal{T}_{a,t}} \left(g(\theta, h_a^{t'}(x_0)) - g(\hat{\theta}_a, h_a^{t'}(\hat{x}_{a,0}))\right)^2 > \gamma,
\end{equation}
contradicting the definition of $\psi_{a,t}(\gamma)$.

Therefore, the assumption is false and we conclude the desired result.
\endproof

\proof{Proof of Proposition \ref{prop2}} 
Motivated by the Proof of Proposition 2 in \cite{l2o}, we consider upper bounds for the two separate terms in Proposition \ref{decom} after the regret decomposition in the following three cases:

\textbf{Case 1:} $\mathbb{E}r_{\pi^*_t,t} > U_t(\pi^*_t)$.

In this case, the expected reward of the optimal policy exceeds its upper confidence bound. We bound the regret directly (prior to decomposition) by the maximum regret $C_g$.

\textbf{Case 2:} $\mathbb{E}r_{\pi^*_t,t} \leq U_t(\pi^*_t)$ and $\mathbb{E}r_{\pi_t,t} > L_t(\pi_t)$.

Here, the first term in the regret decomposition becomes non-positive, then we have:
      \begin{align}
          \mathbb{E}(r_{\pi^*_t,t}-r_{\pi_t,t})&=\mathbb{E} (r_{\pi^*_t,t}-U_t(\pi^*_t))+\mathbb{E}(U_t(\pi_t)-r_{\pi_t,t})\\
          &\leq U_t(\pi_t)-\mathbb{E}r_{\pi_t,t}\\
          &\leq   U_t(\pi_t)-L_t(\pi_t).
      \end{align}

\textbf{Case 3:} $\mathbb{E}r_{\pi^*_t,t} \leq U_t(\pi^*_t)$ and $\mathbb{E}r_{\pi_t,t} \leq L_t(\pi_t)$.

In this case, we also bound the regret directly (prior to decomposition) by the maximum regret $C_g$.
  
Combining the three cases, we obtain the following bound:
\begin{align}
\mathbb{E}(r_{\pi^*_t,t} - r_{\pi_t,t}) 
& \leq C_g \cdot P\left( \mathbb{E}r_{\pi^*_t,t} > U_t(\pi^*_t) \right) + \left( U_t(\pi_t) - L_t(\pi_t) \right) \cdot P\left( \mathbb{E}r_{\pi^*_t,t} \leq U_t(\pi^*_t), \mathbb{E}r_{\pi_t,t} > L_t(\pi_t) \right) \\
&\quad +  C_g \cdot P\left(\mathbb{E}r_{\pi^*_t,t} \leq U_t(\pi^*_t), \mathbb{E}r_{\pi_t,t} \leq L_t(\pi_t) \right)\\
& \leq C_g \cdot P\left( \mathbb{E}r_{\pi^*_t,t} > U_t(\pi^*_t) \right) + \left( U_t - L_t \right)(\pi_t)+C_g\cdot P\left(\mathbb{E}r_{\pi_t,t} \leq L_t(\pi_t) \right)\\
&\leq \sum\limits_{a\in \{\pi_t,\pi^*_t\}} C_g\cdot P\left(\mathbb{E}r_{a,t}\notin [L_t(a), U_t(a)]\right)+ \left( U_t - L_t \right)(\pi_t)\\
&\leq \sum\limits_{a\in\mathcal{A}} C_g\cdot P\left(\mathbb{E}r_{a,t}\notin [L_t(a), U_t(a)]\right)+ \left( U_t - L_t \right)(\pi_t),
\end{align}

then note that the expected reward can be expressed as,
\begin{equation}
    \mathbb{E}r_{a,t} = g(\theta_a^*, h_a^t(x_{a,0}^*)),
\end{equation}

which gives the stated bound.

\endproof

\proof{Proof of lemma \ref{lemma2}}
    Let $t_{a}$ denote the first time that $a$ is chosen in the time horizon $T$. 
We use a narrower confidence interval to demonstrate the high probability of the true reward being within the interval. We define $U_t'(a)$ and $L_t'(a)$ as:

\begin{equation}
    U'_t(a) =
    \begin{cases}
        \max \{g(\theta_a,h_a^{t}(x_{a,0})) :  \frac{1}{2\sigma n(\mathcal{T}_{a,t})}\sum\limits_{t'\in \mathcal{T}_{a,t}}{\left(g(\theta,h_a^{t'}(x_0)) - g(\hat{\theta}_a,h_a^{t'}(\hat{x}_{a,0}))\right)^2} \leq   \gamma_{a,t}\} & \text{if } n(\mathcal{T}_{a,t}) \geq 1, \\
            g_{\max} & \text{if } n(\mathcal{T}_{a,t}) = 0.
    \end{cases}
\end{equation}

\begin{equation}
    L'_t(a) =
    \begin{cases}
        \min \{g(\theta_a,h_a^{t}(x_{a,0})) :  \frac{1}{2\sigma n(\mathcal{T}_{a,t})}\sum\limits_{t'\in \mathcal{T}_{a,t}}{\left(g(\theta,h_a^{t'}(x_0)) - g(\hat{\theta}_a,h_a^{t'}(\hat{x}_{a,0}))\right)^2} \leq   \gamma_{a,t}\} & \text{if } n(\mathcal{T}_{a,t}) \geq 1, \\
            g_{\min} & \text{if } n(\mathcal{T}_{a,t}) = 0.
    \end{cases}
\end{equation}

We consider two cases for each arm $a\in \mathcal{A}$:

 \textbf{Case 1:}   $\forall a\in \mathcal{A}$, when $n(\mathcal{T}_{a,t})=0$, (i.e., $t\in \{1,\dots,T\}$ such that $t< t_a$):
    \begin{equation}\begin{aligned}
    & {P}\left(\{ g(\theta_{a}^*, h_a^{t}(x_{a,0}^*)  )\notin [L'_t(a),U'_t(a)]  \}\right)  ={P}\left(\{ g(\theta_{a}^*, h_a^{t}(x_{a,0}^*)  )\notin [g_{\min},g_{\max}]  \}\right) =0\leq 1/T^2.
    \end{aligned}\end{equation}

   \textbf{Case 2:} $\forall a\in \mathcal{A}$, when $n(\mathcal{T}_{a,t})\geq 1$, (i.e., $t\in \{1,\dots,T\}$ such that $t\geq t_a$):

    Using Theorem \ref{concentration} and let $\alpha=1/T^2$, combined with Proposition \ref{past-reward-kl-divergence}, we have 

\begin{align}
&\quad P\left(\frac{1}{2\sigma n(\mathcal{T}_{a,t})} \sum\limits_{t'\in \mathcal{T}_{a,t}}\left(g(\theta_a^*,h_a^{t'}(x_0^*)) - g(\hat{\theta}_a,h_a^{t'}(\hat{x}_{a,0}))\right)^2> \frac{c_f(d_x,d_\theta)+2L_p\sigma\sqrt{\log(T)}}{\sqrt{n(\mathcal{T}_{a,t})}}\right)\\
&\stackrel{(a)}{\leq} P\left(\frac{1}{n(\mathcal{T}_{a,t})}  D_{a,\pi_1^t}\left(\theta_a^*,x_0^* \| \hat{\theta}_a,\hat{x}_{a,0}\right) > \frac{c_f(d_x,d_\theta)+2L_p\sigma\sqrt{\log(T)}}{\sqrt{n(\mathcal{T}_{a,t})}}\right)\\
&\stackrel{(b)}{\leq} \frac{1}{T^2},
\end{align}
where (a) follows from Proposition \ref{past-reward-kl-divergence} and (b) follows from Theorem \ref{concentration}.

Then the above implies that 
$
{P}\left(\{ g(\theta_{a}^*, h_a^{t}(x_{a,0}^*)  )\notin [L'_t(a),U'_t(a)]  \}\right)\leq 1/T^2
$.

Based on the definition of $\psi_{a,t}$, $U_t(a)$ and $L_t(a)$, we have $U_t(a)\geq U'_t(a)$ and $L_t(a)\leq L'_t(a)$, which implies that:
\begin{equation}\begin{aligned}
    &\sum\limits_{t=1}^T{P}\left(\{ g(\theta_{a}^*, h_a^{t}(x_{a,0}^*)  )\notin [L_t(a),U_t(a)]  \}\right) \leq \sum\limits_{t=1}^T {P}\left(\{ g(\theta_{a}^*, h_a^{t}(x_{a,0}^*)  )\notin [L_t'(a),U_t'(a)]  \}\right) \leq \sum\limits_{t=1}^T \frac{1}{T^2}=\frac{1}{T}.
\end{aligned}\end{equation}
\endproof

\proof{Proof of Theorem \ref{thmReg}}
We consider two cases for the time horizon $T$.

\textbf{Case 1:} When $T\leq|\mathcal{A}|$: 
At most one action can be selected at each time step, and the regret is bounded by the maximum regret $C_g$, leading to:
\begin{equation}
    R_\Pi(T) \leq C_gT=C_g\min\{T,|\mathcal{A}|\}.
\end{equation}

\textbf{Case 2:} When $T>|\mathcal{A}|$:
Using Proposition \ref{decom} and Proposition \ref{prop2}, we have:
{\begin{equation}\begin{aligned}
    R_\Pi(T) &=\sum\limits_{t=1}^T\mathbb{E}(U_t(\pi_t)-r_{\pi_t,t})+\mathbb{E} (r_{\pi^*_t,t}-U_t(\pi^*_t))\\
    &\leq \mathbb{E}\left[\sum_{t=1}^{T}(U_t - L_t)(\pi_t)\right] +\sum\limits_{t=1}^T\sum\limits_{a\in \mathcal{A}} C_g {P}\left( \{g(\theta_{a}^* , h_a^{t}(x_{a,0}^*) )\notin [L_t(a),U_t(a)] \}\right).
  \end{aligned}\end{equation}
}

Then we consider the terms in the above equation separately.

For the first term, we leverage the definitions of  $U_t$ and $L_t$, along with the result in Lemma \ref{lemma1}. Additionally, we exclude the case where $ n(\mathcal{T}_{a,t}) = 0 $,  as control factor $\gamma_{a,t}$ is not well defined in this scenario and we bound the regret by $C_g$ instead. We have:
\begin{align}
    &\mathbb{E}\left[\sum_{t=1}^{T}(U_t - L_t)(\pi_t)\right] = \mathbb{E}\left[\sum_{a\in \mathcal{A}}\sum_{t\in \mathcal{T}_a}(U_t - L_t)(a)\right] \\
    &\leq 2\mathbb{E}\left[\sum_{a\in \mathcal{A} }\sum_{t\in \mathcal{T}_a\setminus\{t_a\} }\psi_{a,t}\left(\frac{c_f(d_x,d_\theta)+2L_p\sigma\sqrt{\log(T)}}{\sqrt{n(\mathcal{T}_{a,t})}}\right)\right]+C_g|\mathcal{A}| \\
    &\stackrel{(a)}{\leq} 2L_g^2 \sum_{a\in \mathcal{A}}\sum_{t\in \mathcal{T}_a\setminus\{t_a\}} \sqrt{\frac{2\sigma c_f(d_x,d_\theta)+4L_p\sigma^2\sqrt{\log(T)}}{\sqrt{n(\mathcal{T}_{a,t})}}}+C_g|\mathcal{A}|\\
    &\stackrel{(b)}{\leq} \frac{8}{3}L_g^2 \sum_{a\in \mathcal{A}}\left(n(\mathcal{T}_a)\right)^{\frac{3}{4}}\sqrt{2\sigma c_f(d_x,d_\theta)+4L_p\sigma^2\sqrt{\log(T)}}+C_g|\mathcal{A}|\\
    &\stackrel{(c)}{\leq} \frac{8}{3} L_g^2|\mathcal{A}|^{\frac{1}{4}}T^{\frac{3}{4}}\sqrt{ 2\sigma c_f(d_x,d_\theta)+4L_p\sigma^2\sqrt{\log(T)}}+C_g|\mathcal{A}|.
\end{align}
where (a) follows from Lemma \ref{lemma1}, $(b)$ follows the below results:
\begin{align}
    \sum_{t \in \mathcal{T}_a \setminus \{ t_a \}} \left( \frac{1}{n(\mathcal{T}_{a,t})} \right)^{1/4} & \leq \int_1^{n(\mathcal{T}_a)} x^{-1/4} \, dx = \frac{4}{3} \left( n(\mathcal{T}_a) \right)^{3/4} - \frac{4}{3} \leq \frac{4}{3} \left( n(\mathcal{T}_a) \right)^{3/4},
\end{align}
with $(c)$ following from Jensen's inequality and the fact that $\sum_{a \in \mathcal{A}} n(\mathcal{T}_a) = T$.

For the second term, we leverage Lemma \ref{lemma2} to show that the true reward is within the confidence intervals with high probability. We have:
\begin{equation}
\begin{aligned}
    \sum_{t=1}^T \sum_{a \in \mathcal{A}} C_g  \, 
    P\big( g(\theta_a^*, h_a^{t}(x_{a,0}^*)) \notin [L_t(a), U_t(a)] \big)
    \leq \sum_{a \in \mathcal{A}} \frac{C_g}{T} \leq C_g |\mathcal{A}|.
\end{aligned}
\end{equation}

Summing up the two terms, we have:
\begin{equation}\begin{aligned}
    R_\Pi(T) &\leq \frac{8}{3}L_g^2 |\mathcal{A}|^{\frac{1}{4}}T^{\frac{3}{4}}\sqrt{ 2\sigma c_f(d_x,d_\theta) + 4L_p\sigma^2\sqrt{\log(T)}} +2C_g\min\{T,|\mathcal{A}|\},
\end{aligned}\end{equation}
which completes the proof.
\endproof

\begin{theorem}[Lindeberg–Feller Central Limit Theorem \cite{van2000asymptotic}]
\label{LF-CLT}
Let $\{X_i\}_{i=1}^n$ be a sequence of independent, mean-zero random variables with variances $\sigma_i^2 = \mathbb{E}[X_i^2]$, and total variance
$s_n^2 = \sum_{i=1}^n \sigma_i^2$. Suppose that for every $\varepsilon > 0$, the Lindeberg condition holds:
\begin{equation}
\frac{1}{s_n^2} \sum_{i=1}^n \mathbb{E}\big[ X_i^2 \mathbf{1}{\{|X_i| > \varepsilon s_n\}} \big] \to 0 \quad \text{as } n \to \infty.
\end{equation}
Then the normalized sum
\begin{equation}
\frac{S_n}{s_n} = \frac{\sum_{i=1}^n X_i}{s_n}
\end{equation}
converges in distribution to a standard normal random variable:
\begin{equation}
\frac{S_n}{s_n} \xrightarrow{d} \mathcal{N}(0,1) \quad \text{as } n \to \infty.
\end{equation}
\end{theorem}

\begin{proposition}
\label{lindberg}
Let $\{X_i\}_{i=1}^n$ be a sequence of independent, mean-zero sub-Gaussian random variables with variances $\sigma_i^2 = \mathbb{E}[X_i^2]<\infty$. Assume there exists constant $\sigma$ such that $\sigma_i^2 \geq \sigma^2$ for all $i$, and that there exists a constant $K > 0$ such that each $X_i$ is sub-Gaussian with parameter at most $K$, i.e.,
\begin{equation}
\label{subg}
{P}(|X_i| > t) \leq 2 \exp\left( -\frac{t^2}{2K^2} \right) \quad \text{for all } t > 0.
\end{equation}
Then the Lindeberg condition is satisfied.
\end{proposition}

\proof{Proof of Proposition \ref{lindberg}}

Fix $\varepsilon > 0$. The Lindeberg condition requires
\begin{equation}
\frac{1}{s_n^2} \sum_{i=1}^n \mathbb{E}\left[ X_i^2 \mathbf{1}{\{|X_i| > \varepsilon s_n\}} \right] \to 0 \quad \text{as } n \to \infty.
\end{equation}

Using integration by tails,
\begin{equation}
\begin{aligned}
\mathbb{E}\left[ X_i^2 \mathbf{1}{\{|X_i| > \varepsilon s_n\}} \right] 
&= \int_{\varepsilon^2 s_n^2}^\infty {P}(X_i^2 > u) \, du
= \int_{\varepsilon s_n}^\infty {P}(|X_i| > t) \cdot 2t \, dt \\
&\stackrel{(a)}{\leq} \int_{\varepsilon s_n}^\infty 4t \exp\left(-\frac{t^2}{2K^2}\right) dt
= 4K^2 \int_{\frac{\varepsilon^2 s_n^2}{2K^2}}^\infty e^{-u} du
= 4K^2 \exp\left(-\frac{\varepsilon^2 s_n^2}{2K^2}\right),
\end{aligned}
\end{equation}
where (a) follows from the sub-Gaussian tail bound \eqref{subg}.

Summing and normalizing,
\begin{equation}
\frac{1}{s_n^2} \sum_{i=1}^n \mathbb{E}\left[ X_i^2 \mathbf{1}_{\{|X_i| > \varepsilon s_n\}} \right]
\leq \frac{4nK^2}{s_n^2} \exp\left(-\frac{\varepsilon^2 s_n^2}{2K^2}\right).
\end{equation}

Now observe that since $\sigma_i^2 \geq \sigma^2 > 0$ for all $i$, 
\begin{equation}
\frac{n}{s_n^2} \leq \frac{1}{\sigma^2} \quad \text{for all } n.
\end{equation}

Therefore,
\begin{equation}
\frac{1}{s_n^2} \sum_{i=1}^n \mathbb{E} \left[ X_i^2 \mathbf{1}_{\{ |X_i| > \varepsilon s_n \}} \right]
\leq \frac{4K^2}{\sigma^2} \exp\left( -\frac{\varepsilon^2 \sigma^2n}{2K^2} \right) \to 0 \quad \text{as } n \to \infty,
\end{equation}
which verifies the Lindeberg condition.
\endproof

\proof{Proof of Theorem \ref{individual-MLE}}
For a given user $i$ and arm $a$, the system dynamics are given by:
\begin{equation}
    x_{i,a,t+1} = A x_{i,a,t} + B \pi_{i,a,t} + K
\end{equation}
By unrolling this recurrence relation, we can find a closed-form expression for the state $x_{i,a,t}$:
\begin{equation}
    x_{i,a,t} = A^t x_{i,a,0} + \sum_{j=0}^{t-1} A^{t-1-j} (B \pi_{i,a,j} + K)
\end{equation}

Also the reward can be expressed as is $r_{i,a,t} = C x_{i,a,t} + D \theta_a + \varepsilon_{i,a,t}$, where $\varepsilon_{i,a,t}$ are i.i.d. mean-zero sub-Gaussian noise terms with variance $\sigma^2$ and and sub-Gaussian parameter $\overline{\sigma}$.

Substituting the expression for $x_{i,a,t}$ into the reward model gives:
\begin{equation}
    r_{i,a,t} = C \left( A^t x_{i,a,0} + \sum_{j=0}^{t-1} A^{t-1-j} (B \pi_{i,a,j} + K) \right) + D \theta_a + \varepsilon_{i,a,t}
\end{equation}
We can rearrange this equation to separate the terms containing the unknown parameters ($x_{i,a,0}, \theta_a$) from the known terms. Let $K'_{i,a,t} = C \sum_{j=0}^{t-1} A^{t-1-j} (B \pi_{i,a,j} + K)$ be the part of the reward determined by the known system parameters and historical actions.
 Let the parameter vector be $z_{i,a} = [x_{i,a,0}, \theta_a]^\top$ and the design vector at time $t$ be $\phi_t = [C A^t, D]^\top$. Let the modified reward be $r'_{i,a,t} = r_{i,a,t} - K'_{i,a,t}$. The model is then: $r'_{i,a,t} = \phi_t^\top z_{i,a} + \varepsilon_{i,a,t}$.

The least-squares estimator $\hat{z}_{i,a}$ minimizes the sum of squared errors, $S(z_{i,a})$. Since $K'_{i,a,t}$ does not depend on $z_{i,a}$, minimizing the sum of squared errors for $r_{i,a,t}$ is equivalent to minimizing it for $r'_{i,a,t}$:
\begin{align}
    S(z_{i,a}) &=\sum_{t \in \mathcal{T}_{i,a}} (r_{i,a,t} - \phi_t^\top z_{i,a} -K'_{i,a,t})^2  \\
    &\sum_{t \in \mathcal{T}_{i,a}} (r'_{i,a,t} - \phi_t^\top z_{i,a})^2
\end{align}
To find the minimum, we take the gradient with respect to $z_{i,a}$ and set it to zero:
\begin{align}
    \nabla_{z_{i,a}} S(z_{i,a}) &= \sum_{t \in \mathcal{T}_{i,a}} 2(r'_{i,a,t} - \phi_t^\top z_{i,a})(-\phi_t) = 0 \\
    \implies \sum_{t \in \mathcal{T}_{i,a}} r'_{i,a,t}\phi_t &= \left(\sum_{t \in \mathcal{T}_{i,a}} \phi_t \phi_t^\top \right) z_{i,a}
\end{align}
Solving for $z_{i,a}$ gives the solution, $\hat{z}_{i,a}$:
\begin{equation}
    \hat{z}_{i,a} = \left( \sum_{t \in \mathcal{T}_{i,a}} \phi_t \phi_t^\top \right)^{-1} \left( \sum_{t \in \mathcal{T}_{i,a}} \phi_t r'_{i,a,t} \right)
\end{equation}

From the theorem, the design matrix is $\mathcal{I}_{i,a}(z_{i,a}^*) = \frac{1}{\sigma^2} \sum_{t \in \mathcal{T}_{i,a}} 
 \begin{bmatrix} 
 (C A^{t})^2 & C A^{t} D \\ 
 C A^{t} D & D^2 
 \end{bmatrix}$.
We can see that $\phi_t \phi_t^\top = \begin{bmatrix} C A^t \\ D \end{bmatrix} \begin{bmatrix} C A^t & D \end{bmatrix} = \begin{bmatrix} (C A^t)^2 & C A^t D \\ C A^t D & D^2 \end{bmatrix}$.
Therefore, $\sum_{t \in \mathcal{T}_{i,a}} \phi_t \phi_t^\top = \sigma^2 \mathcal{I}_{i,a}(z_{i,a}^*)$. Let's denote $\mathcal{I}_{i,a} = \mathcal{I}_{i,a}(z_{i,a}^*)$ for brevity.

Now, substitute the true model $r'_{i,a,t} = \phi_t^\top z_{i,a}^* + \varepsilon_{i,a,t}$ (where $z_{i,a}^*$ is the true parameter vector) into the estimator equation:
\begin{align*}
    \hat{z}_{i,a} &= (\sigma^2 \mathcal{I}_{i,a})^{-1} \sum_{t \in \mathcal{T}_{i,a}} \phi_t (\phi_t^\top z_{i,a}^* + \varepsilon_{i,a,t}) \\
    &= (\sigma^2 \mathcal{I}_{i,a})^{-1} \left( \sum_{t \in \mathcal{T}_{i,a}} \phi_t \phi_t^\top \right) z_{i,a}^* + (\sigma^2 \mathcal{I}_{i,a})^{-1} \sum_{t \in \mathcal{T}_{i,a}} \phi_t \varepsilon_{i,a,t} \\
    &= z_{i,a}^* + (\sigma^2 \mathcal{I}_{i,a})^{-1} \sum_{t \in \mathcal{T}_{i,a}} \phi_t \varepsilon_{i,a,t}
\end{align*}

The term $\sum_{t \in \mathcal{T}_{i,a}} \phi_t \varepsilon_{i,a,t}$ is a weighted sum of i.i.d. sub-Gaussian random variables. A linear transformation of sub-Gaussian variables is also sub-Gaussian. Thus, $\hat{z}_{i,a}$ is also sub-Gaussian distributed.

The mean of the estimator is:
\begin{equation}
    \mathbb{E}[\hat{z}_{i,a}] = \mathbb{E}\left[z_{i,a}^* + (\sigma^2 \mathcal{I}_{i,a})^{-1} \sum_{t \in \mathcal{T}_{i,a}} \phi_t \varepsilon_{i,a,t}\right] = z_{i,a}^*
\end{equation}
since $\mathbb{E}[\varepsilon_{i,a,t}] = 0$. The estimator is unbiased.

The covariance of the estimator is:
\begin{align}
    \cov(\hat{z}_{i,a}) &= \cov\left( (\sigma^2 \mathcal{I}_{i,a})^{-1} \sum_{t \in \mathcal{T}_{i,a}} \phi_t \varepsilon_{i,a,t} \right) \\
    &= (\sigma^2 \mathcal{I}_{i,a})^{-1} \left( \sum_{t \in \mathcal{T}_{i,a}} \phi_t \var(\varepsilon_{i,a,t}) \phi_t^\top \right) ((\sigma^2 \mathcal{I}_{i,a})^{-1})^\top \\
    &= (\sigma^2 \mathcal{I}_{i,a})^{-1} \left( \sigma^2 \sum_{t \in \mathcal{T}_{i,a}} \phi_t \phi_t^\top \right) (\sigma^2 \mathcal{I}_{i,a})^{-1} \\
    &= (\sigma^2 \mathcal{I}_{i,a})^{-1} (\sigma^2 (\sigma^2 \mathcal{I}_{i,a})) (\sigma^2 \mathcal{I}_{i,a})^{-1} = \mathcal{I}_{i,a}^{-1}
\end{align}
Thus, we have shown that $\hat{z}_{i,a}$ is sub-Gaussian with mean $z_{i,a}^*$ and covariance $\mathcal{I}_{i,a}^{-1}$.

The marginal distribution for the component $\hat{\theta}_{i,a}$ is also sub-Gaussian. Its mean is the corresponding element of $z_{i,a}^*$, which is $\theta_a^*$. Its variance is the corresponding diagonal element of the covariance matrix $\mathcal{I}_{i,a}^{-1}$, which is $[\mathcal{I}_{i,a}^{-1}]_{2,2}$. 
For the estimator to be well-defined, the matrix $\mathcal{I}_{i,a}$ must be invertible. Its determinant is given by:
\begin{equation}
    \det(\mathcal{I}_{i,a}) = \frac{D^2}{\sigma^4} \left( |\mathcal{T}_{i,a}| \sum_{t \in \mathcal{T}_{i,a}} (C A^t)^2 - \left( \sum_{t \in \mathcal{T}_{i,a}} C A^t \right)^2 \right)
\end{equation}
The determinant is nonzero unless $C A^t$ is constant for all $t \in \mathcal{T}_{i,a}$, by the Cauchy–Schwarz inequality. We assumed $A \neq 1$, then this implies that the arm is observed at least at two time steps ($|\mathcal{T}_{i,a}| \geq 2$) for $\mathcal{I}_{i,a}$ to be invertible.

To compute the variance of $\hat{\theta}_{i,a}$, we explicitly invert the design matrix:
\begin{align}
    \mathcal{I}_{i,a}^{-1} 
    &=  \frac{\sigma^2}{D^2 \left( |\mathcal{T}_{i,a}| \sum\limits_{t \in \mathcal{T}_{i,a}} (C A^t)^2 
    - \left( \sum\limits_{t \in \mathcal{T}_{i,a}} C A^t \right)^2 \right)} \notag \\
    &\quad \cdot 
    \begin{bmatrix}
        D^2 |\mathcal{T}_{i,a}| & -D \sum\limits_{t \in \mathcal{T}_{i,a}} C A^t \\
        -D \sum\limits_{t \in \mathcal{T}_{i,a}} C A^t & \sum\limits_{t \in \mathcal{T}_{i,a}} (C A^t)^2
    \end{bmatrix}
\end{align}

The variance of $\hat{\theta}_{i,a}$ corresponds to the $(2,2)$-entry of the covariance matrix $\mathcal{I}_{i,a}^{-1}$:
\begin{equation}
    \var(\hat{\theta}_{i,a}) = \frac{\sigma^2}{D^2 \left(
    |\mathcal{T}_{i,a}| 
    - \dfrac{\left( \sum_{t \in \mathcal{T}_{i,a}} A^t \right)^2}{\sum_{t \in \mathcal{T}_{i,a}} (A^t)^2}
  \right)}.
\end{equation}

Note that since the time horizon $T$ is fixed, there are only finitely many possible subsets $\mathcal{T}_{i,a} \subseteq {1, \dots, T}$. Consequently, for all subsets with $|\mathcal{T}_{i,a}| \ge 2$, the variance $\var(\hat{\theta}_{i,a})$ attains uniform positive lower and upper bounds over all $i$ for any fixed $a$. We denote the minimum and maximum bounds by $c$ and $C$, respectively:
\begin{equation}
\label{var-limit}
c \le \var(\hat{\theta}_{i,a}) \le C.
\end{equation}

Then we investigate the sub-Gaussian parameter of individual estimators. We define the unit vector $e_2=[0,1]^\top$, then we have 
\begin{equation}
    \hat\theta_{i,a}-\theta_a^*=e_2^\top(\hat{z}_{i,a}-z_{i,a}^*)=e_2^\top \left( \sum_{t \in \mathcal{T}_{i,a}} \phi_t \phi_t^\top \right)^{-1} \sum_{t \in \mathcal{T}_{i,a}} \phi_t \varepsilon_{i,a,t}.
\end{equation}

This shows that the estimation error is a linear combination of the independent, zero-mean, $\overline{\sigma}$-sub-Gaussian noise terms $\varepsilon_{i,a,t}$. We can write this as:
\begin{equation}
    \hat{\theta}_{i,a} - \theta_a^* = \sum_{t \in \mathcal{T}_{i,a}} w_t \varepsilon_{i,a,t},
\end{equation}
where the weights $w_t$ are defined as $w_t = e_2^\top \left( \sum_{j \in \mathcal{T}_{i,a}} \phi_j \phi_j^\top \right)^{-1} \phi_t$.

The squared sub-Gaussian parameter of $\hat{\theta}_{i,a} - \theta_a^*$, denoted as $\nu_{i,a}^2$, is given by the sum of squared weights multiplied by the noise's squared sub-Gaussian parameter:
\begin{equation}
    \nu_{i,a}^2 = \overline{\sigma}^2 \sum_{t \in \mathcal{T}_{i,a}} w_t^2.
\end{equation}

Then we calculate the exact value:
\begin{align}
 \sum_{t \in \mathcal{T}_{i,a}} w_t^2 &= \sum_{t \in \mathcal{T}_{i,a}} \left(e_2^\top \left(\sum_{j \in \mathcal{T}_{i,a}} \phi_j \phi_j^\top\right)^{-1} \phi_t\right) \left(\phi_t^\top \left(\left(\sum_{j \in \mathcal{T}_{i,a}} \phi_j \phi_j^\top\right)^{-1}\right)^\top e_2\right) \\
 &= e_2^\top \left(\sum_{j \in \mathcal{T}_{i,a}} \phi_j \phi_j^\top\right)^{-1} \left( \sum_{t \in \mathcal{T}_{i,a}} \phi_t \phi_t^\top \right) \left(\left(\sum_{j \in \mathcal{T}_{i,a}} \phi_j \phi_j^\top\right)^{-1}\right)^\top e_2\\
 &= e_2^\top \left(\sum_{t \in \mathcal{T}_{i,a}} \phi_t \phi_t^\top\right)^{-1} \left( \sum_{t \in \mathcal{T}_{i,a}} \phi_t \phi_t^\top \right) \left(\sum_{t \in \mathcal{T}_{i,a}} \phi_t \phi_t^\top\right)^{-1} e_2 \\
 &= e_2^\top \left( \sum_{t \in \mathcal{T}_{i,a}} \phi_t \phi_t^\top \right)^{-1} e_2
\end{align}

Then note that $\sum_{t \in \mathcal{T}_{i,a}} \phi_t \phi_t^\top = \sigma^2 \mathcal{I}_{i,a}(z_{i,a}^*)$,
and $
    \var(\hat{\theta}_{i,a}) = e_2^\top \cov(\hat{z}_{i,a}) e_2 = \sigma^2 e_2^\top \left( \sum_{t \in \mathcal{T}_{i,a}} \phi_t \phi_t^\top \right)^{-1} e_2
$, therefore, we have \begin{equation}
    \sum_{t \in \mathcal{T}_{i,a}} w_t^2 = \frac{\var(\hat{\theta}_{i,a})}{\sigma^2}
\end{equation}
Finally, substituting this back into the formula for $\nu_{i,a}^2$:
\begin{equation}
    \nu_{i,a}^2 = \overline{\sigma}^2 \left( \frac{\var(\hat{\theta}_{i,a})}{\sigma^2} \right) = \frac{\overline{\sigma}^2}{\sigma^2} \var(\hat{\theta}_{i,a})
\end{equation}
Thus, the estimator $\hat{\theta}_{i,a}$ is sub-Gaussian with mean $\theta_a^*$ and sub-Gaussian parameter $\nu_{i,a}$ given by and upper bounded using Equation \ref{var-limit}:
\begin{equation}
\label{sub-para}
    \nu_{i,a} = \sqrt{\frac{\overline{\sigma}^2}{\sigma^2} \var(\hat{\theta}_{i,a})} = \frac{\overline{\sigma}}{\sigma} \sqrt{\var(\hat{\theta}_{i,a})}\leq \frac{\overline{\sigma}}{\sigma} \sqrt{C}
\end{equation}

We now verify the conditions to apply the Lindeberg-Feller CLT to the sequence of centered estimators $X_i = \hat{\theta}_{i,a} - \theta_a^*$.
The estimators for each user are assumed to be independent under Assumption \ref{extendDependence}. Each estimator $\hat\theta_{i,a}$ has mean $\theta_a^*$, meaning that the expected value of $X_i$ is zero. Additionally, the variance of each $X_i$ is uniformly bounded below by a positive constant $c > 0$. Finally, each $X_i$ is sub-Gaussian with a parameter that is uniformly bounded across all users by $\frac{\overline{\sigma}}{\sigma} \sqrt{C}$.
This sequence satisfies all premises of Proposition \ref{lindberg}, which implies the Lindeberg condition holds. We can therefore invoke the Lindeberg–Feller CLT (Theorem \ref{LF-CLT}). Let $S_N = \sum_{i=1}^N X_i$ and $s_N^2 = \sum_{i=1}^N \var(\hat{\theta}_{i,a})$. The CLT gives:
\begin{equation*}
\frac{S_N}{s_N} = \frac{\sum_{i=1}^N (\hat{\theta}_{i,a} - \theta_a^*)}{ \left( \sum_{i=1}^N \var(\hat{\theta}_{i,a}) \right)^{1/2} } \xrightarrow{d} \mathcal{N}(0,1) \quad \text{as } N \to \infty.
\end{equation*}
This convergence in distribution implies that for large $N$, the sum $S_N = N(\hat{\theta}_a - \theta_a^*)$ is approximately normal. Scaling this result gives the asymptotic distribution of the pooled estimator:
 \begin{equation}
        \hat{\theta}_a \approx \mathcal{N} \left(
        \theta_a^*,\;
        \left( \frac{1}{N^2} \sum_{i=1}^N \var(\hat{\theta}_{i,a})\right)^{1/2}
        \right).
    \end{equation}
\endproof

\proof{Proof of Proposition \ref{fisherBound}}
\textbf{Case 1}: When $A\in (0,1)$.

We begin by expanding the square in the numerator:
\begin{equation}
\left( \sum_{t \in \mathcal{T}} A^t \right)^2 = \sum_{t \in \mathcal{T}} A^{2t} + 2 \sum_{\substack{s, t \in \mathcal{T} \\ s < t}} A^{s + t}.
\end{equation}

Therefore,
\begin{equation}
\frac{\left( \sum_{t \in \mathcal{T}} A^t \right)^2}{\sum_{t \in \mathcal{T}} A^{2t}} = 1 + \frac{2 \sum_{\substack{s < t \\ s, t \in \mathcal{T}}} A^{s + t}}{\sum_{t \in \mathcal{T}} A^{2t}}.
\end{equation}

Then note that:
\begin{equation}
\sum_{\substack{s < t \\ s, t \in \mathcal{T}}} A^{s + t} = \sum_{s \in \mathcal{T}} A^s \sum_{\substack{t \in \mathcal{T} \\ t > s}} A^t.
\end{equation}

Since $t > s$, the inner sum can be bounded by a geometric series:
\begin{equation}
\sum_{\substack{t \in \mathcal{T} \\ t > s}} A^t \leq \sum_{k = s + 1}^\infty A^k = \frac{A^{s + 1}}{1-A}.
\end{equation}

Thus,
\begin{equation}
\sum_{s \in \mathcal{T}} A^s \sum_{\substack{t \in \mathcal{T} \\ t > s}} A^t \leq \sum_{s \in \mathcal{T}} A^s \cdot \frac{A^{s + 1}}{1-A} = \frac{A}{1-A} \sum_{s \in \mathcal{T}} A^{2s}.
\end{equation}

So we have:
\begin{equation}
\sum_{\substack{s < t \\ s, t \in \mathcal{T}}} A^{s + t} \leq \frac{A}{1-A} \sum_{t \in \mathcal{T}} A^{2t}.
\end{equation}

Therefore,
\begin{equation}
\frac{\left( \sum_{t \in \mathcal{T}} A^t \right)^2}{\sum_{t \in \mathcal{T}} A^{2t}} \leq 1 + \frac{2A}{1-A} = \frac{1 + A}{1-A} =\frac{1 + |A|}{|1 - |A||}.
\end{equation}

\textbf{Case 2}: When $A\in (1,+\infty)$.

Following the same initial step as in Case 1, we start with the expanded expression:
\begin{equation}
\frac{\left( \sum_{t \in \mathcal{T}} A^t \right)^2}{\sum_{t \in \mathcal{T}} A^{2t}} = 1 + \frac{2 \sum_{\substack{s < t \\ s, t \in \mathcal{T}}} A^{s + t}}{\sum_{t \in \mathcal{T}} A^{2t}}.
\end{equation}

Then note that:
\begin{equation}
\sum_{\substack{s < t \\ s, t \in \mathcal{T}}} A^{s + t} = \sum_{t \in \mathcal{T}} A^t \sum_{\substack{s \in \mathcal{T} \\ s<t}} A^s.
\end{equation}

Since $ t > s $, the inner sum can be bounded by a geometric series:
\begin{align}
\sum_{\substack{s \in \mathcal{T} \\ s < t}} A^s 
&< \sum_{k=0}^{\infty} A^{t-1-k} = A^{t-1} \sum_{k=0}^{\infty} \left(\frac{1}{A}\right)^k \\
&= A^{t-1} \left( \frac{A}{A - 1} \right) = \frac{A^t}{A - 1}.
\end{align}

Thus, 
\begin{equation}
\sum_{\substack{s < t \\ s, t \in \mathcal{T}}} A^{s + t} < \sum_{t \in \mathcal{T}} A^t \left( \frac{A^t}{A-1} \right) = \frac{1}{A-1} \sum_{t \in \mathcal{T}} A^{2t}.
\end{equation}

Therefore,
\begin{equation}
\frac{\left( \sum_{t \in \mathcal{T}} A^t \right)^2}{\sum_{t \in \mathcal{T}} A^{2t}} \leq 
1 + \frac{2}{A-1} = \frac{A+1}{A-1}=\frac{1 + |A|}{|1 - |A||}.
\end{equation}

\textbf{Case 3}: When $A\in (-\infty,0)$ and $A\neq -1$.

Let $B = |A|$. Then note that:
\begin{equation}
\frac{\left(\sum_{t \in \mathcal{T}} A^t\right)^2}{\sum_{t \in \mathcal{T}} A^{2t}} \leq \frac{\left(\sum_{t \in \mathcal{T}} |A|^t\right)^2}{\sum_{t \in \mathcal{T}} |A|^{2t}} = \frac{\left(\sum_{t \in \mathcal{T}} B^t\right)^2}{\sum_{t \in \mathcal{T}} B^{2t}}.
\end{equation}
Since $B = |A| > 0$ and $B \neq 1$, this new expression is covered by \textbf{Case 1} (if $0 < B < 1$) or \textbf{Case 2} (if $B > 1$). In both scenarios, we established that:
\begin{equation}
\frac{\left(\sum_{t \in \mathcal{T}} B^t\right)^2}{\sum_{t \in \mathcal{T}} B^{2t}} \leq \frac{1+B}{|1-B|}.
\end{equation}
Substituting $B = |A|$ gives the desired result:
\begin{equation}
\frac{\left(\sum_{t \in \mathcal{T}} A^t\right)^2}{\sum_{t \in \mathcal{T}} A^{2t}} \leq \frac{1+|A|}{|1-|A||}.
\end{equation}

This completes the proof.
\endproof

\proof{Proof of Proposition \ref{betaCalculation}}  
We analyze the Type I and Type II errors under the given hypothesis testing framework.

\textbf{Type I Error:}

Under the null hypothesis $H_0$, the estimators and their variances are given by:
\begin{equation}
\begin{aligned}
\hat\theta_a &= \frac{1}{N} \sum_{i=1}^N \hat{\theta}_{i,a}, \quad
\var(\hat\theta_a) = \frac{1}{N^2} \sum_{i=1}^N \left(\sum\limits_{t \in \mathcal{T}_{i,a}} \mathcal{I}_{i,T}(\theta_a^*)\right)^{-1}, \\
\hat\theta_{a'} &= \frac{1}{N} \sum_{i=1}^N \hat{\theta}_{i,a'}, \quad
\var(\hat\theta_{a'}) = \frac{1}{N^2} \sum_{i=1}^N \left(\sum\limits_{t \in \mathcal{T}_{a',i}} \mathcal{I}_{i,t}(\theta_{a'})\right)^{-1}.
\end{aligned}
\end{equation}

To test for a difference between arms $a$ and $a'$, we construct the test statistic:
\begin{equation}
\begin{aligned}
z &= \frac{\hat{\theta}_a - \hat{\theta}_{a'}}{\sqrt{\var(\hat{\theta}_a) + \var(\hat{\theta}_{a'})}}, \quad z \sim \mathcal{N}(0, 1), \\
P & (z \geq z_\alpha \mid H_0) = \alpha, \quad z_\alpha = \Phi^{-1}(1 - \alpha),
\end{aligned}
\end{equation}
where $\Phi(\cdot)$ denotes the cumulative distribution function of the standard normal distribution.

\textbf{Type II Error:}

Under the alternative hypothesis $H_1$, we assume the true mean difference to be $\Delta$. The non-centrality parameter is:
\begin{equation}
\begin{aligned}
\theta_\Delta &= \frac{\theta_{a} - \theta_{a'}}{\sqrt{\var(\hat{\theta}_a) + \var(\hat{\theta}_{a'})}} = \frac{\Delta}{\sqrt{\var(\hat{\theta}_a) + \var(\hat{\theta}_{a'})}}, \\
\beta &= P(z < z_\alpha \mid H_1),
\end{aligned}
\end{equation}
where under $H_1$, the statistic $z \sim \mathcal{N}(\theta_\Delta, 1)$.

Therefore, the Type II error can be computed as:
\begin{equation}
\beta = \Phi(z_\alpha - \theta_\Delta).
\end{equation}

Substituting the expressions for $\theta_\Delta$ and $z_\alpha$, we obtain:
\begin{equation}
\beta = \Phi\left(\Phi^{-1}(1 - \alpha) - \frac{\Delta}{\sqrt{\var(\hat{\theta}_a) + \var(\hat{\theta}_{a'})}}\right).
\end{equation}
\endproof

\proof{Proof of Theorem \ref{piMin}}  
Assume that the desired Type I error, Type II error, and minimum detectable mean shift are denoted by $\alpha_0$, $\beta_0$, and $\Delta_0$, respectively, in the context of multiple comparisons.

We consider comparisons between arm $a'$ and all other arms $a \in \mathcal{A} \setminus \{a'\}$, resulting in $|\mathcal{A}| - 1$ pairwise tests. To control the family-wise type I and type II error rates, denoted by $\bar{\alpha}$ and $\bar{\beta}$ respectively, we apply Bonferroni correction \citep{holm1979simple} and enforce the following constraints:
\begin{equation}
\begin{aligned}
    \bar{\alpha} &\leq (|\mathcal{A}| - 1)\alpha \leq \alpha_0, \\
    \bar{\beta} &\leq (|\mathcal{A}| - 1)\beta \leq \beta_0,
\end{aligned}
\end{equation}
where $\alpha$ and $\beta$ are the per-comparison type I and type II error rates. These constraints ensure that each comparison is conducted at a sufficiently small error level so that the overall family-wise error rates remain below the predefined thresholds $\alpha_0$ and $\beta_0$. 

Referring to Proposition \ref{betaCalculation}, this leads to the following bound on $\beta$:
\begin{equation}
\beta = \Phi\left(\Phi^{-1}\left(1 - \frac{\alpha_0}{|\mathcal{A}| - 1}\right) - \frac{\Delta}{\sqrt{\var(\hat{\theta}_a) + \var(\hat{\theta}_{a'})}}\right) \leq \frac{\beta_0}{|\mathcal{A}| - 1}.
\end{equation}

To satisfy this inequality, the sum of variances must be sufficiently small:
\begin{equation}
\var(\hat{\theta}_a) + \var(\hat{\theta}_{a'}) \leq \frac{\Delta_0^2}{\left( \Phi^{-1}\left(1 - \frac{\alpha_0}{|\mathcal{A}| - 1} \right) - \Phi^{-1}\left( \frac{\beta_0}{|\mathcal{A}| - 1} \right) \right)^2}.
\end{equation}

Let $p_{\min}$ denote the minimum probability of selecting any arm at each trial. We now derive a lower bound for $p_{\min}$ using an upper bound on the variances. Observe that:
\begin{equation}
\begin{aligned}
\var(\hat\theta_a) 
&= \frac{1}{N^2} \sum_{i=1}^N \var(\hat\theta_{i,a}) \\
&\leq \frac{1}{N^2} \sum_{i=1}^N 
\frac{\sigma^2}{D^2 \left(
    |\mathcal{T}_{i,a}| 
    - \frac{1 + |A|}{|1 - |A||}
\right)}.
\end{aligned}
\end{equation}

Similarly, we have:
\begin{equation}
\var(\hat\theta_{a'}) \leq \frac{1}{N^2} \sum_{i=1}^N 
\frac{\sigma^2}{D^2 \left(
    |\mathcal{T}_{i,a'}| 
    - \frac{1 + |A|}{|1 - |A||}
\right)}.
\end{equation}

To ensure the sum of variances is small enough, all arms must be sufficiently explored.

Then the total variance is bounded as:
\begin{equation}
\begin{aligned}
\var(\hat{\theta}_a) + \var(\hat{\theta}_{a'}) 
&\leq \frac{1}{N^2} \sum_{i=1}^N 
\frac{\sigma^2}{D^2 \left(
    |\mathcal{T}_{i,a}| 
    - \frac{1 + |A|}{|1 - |A||}
\right)} + \frac{1}{N^2} \sum_{i=1}^N 
\frac{\sigma^2}{D^2 \left(
    |\mathcal{T}_{i,a'}| 
    - \frac{1 + |A|}{|1 - |A||}
\right)}\\
&\leq \frac{2\sigma^2}{ND^2 \left(
    Tp_{\min} 
    - \frac{1 + |A|}{|1 - |A||}
\right)} \\
&\leq \frac{\Delta_0^2}{\left( \Phi^{-1}\left(1 - \frac{\alpha_0}{|\mathcal{A}| - 1} \right) - \Phi^{-1}\left( \frac{\beta_0}{|\mathcal{A}| - 1} \right) \right)^2}.
\end{aligned}
\end{equation}

Solving for $p_{\min}$ yields the desired lower bound:
\begin{equation}
p_{\min} \geq \frac{1}{T} \left( \frac{1 + |A|}{|1 - |A||} + \frac{2\sigma^2 \left( \Phi^{-1}\left(1 - \frac{\alpha_0}{|\mathcal{A}| - 1} \right) - \Phi^{-1}\left( \frac{\beta_0}{|\mathcal{A}| - 1} \right) \right)^2}{ND^2\Delta_0^2} \right).
\end{equation}

Finally, to ensure feasibility in exploring all arms, the maximum selection probability must satisfy: $p_{\max} \leq 1 - (|\mathcal{A}| - 1)p_{\min}$.
\endproof

\proof{Proof of Theorem \ref{clipReg}}
The proof follows a similar argument to that in \citet{yao2021power}. Consider a policy $\Pi$ that achieves a regret rate $R_{\Pi}(T)$, and let $\Pi'$ be the clipped version of $\Pi$, obtained by applying the clipping procedure described in Definition~\ref{clipProcess}, such that all action probabilities are lower bounded by $p_{\min} > 0$.

In each round, clipping may alter the action selected by $\Pi$. In the worst case, when $\Pi$ is deterministic and selects a single action with probability one, the clipped distribution assigns probability $p_{\min}$ to each of the remaining $|\mathcal{A}| - 1$ actions. Therefore, the probability that $\Pi'$ selects a different action than $\Pi$ is at most $(|\mathcal{A}| - 1) p_{\min}$. As we defined earlier, $p_{\max} = 1 - (|\mathcal{A}| - 1) p_{\min}$, so $\Pi'$ selects the same action as $\Pi$ with probability at least $p_{\max}$.
Over $T$ rounds, the expected number of rounds on which the policies disagree is at most $(1 - p_{\max}) T$, and the number of rounds on which they agree is at least $p_{\max} T$. On rounds where $\Pi'$ agrees with $\Pi$, the regret incurred is identical.

Here we add an important clarification: even in rounds where $\Pi'$ takes a different action than $\Pi$, we still feed the observed feedback back to $\Pi$. This ensures that $\Pi$ continues to update its internal state (e.g., posterior in ROGUE-TS) based on all observed data, regardless of whether its chosen action was actually executed.

Therefore, although $\Pi$ receives full feedback and keeps learning over all $T$ rounds, it effectively only participates in decision making on approximately $p_{\max} T$ rounds, the rounds where its action is actually taken by $\Pi'$. From this perspective, it is natural to bound the regret contribution of $\Pi$ by $R_{\Pi}(p_{\max} T)$.
On the remaining $(1 - p_{\max}) T$ rounds, where $\Pi'$ deviates from $\Pi$, we conservatively bound the per-step regret by $C_g$. Hence, the total regret of $\Pi'$ is bounded by:
$R_{\Pi'}(T) \leq R_{\Pi}(p_{\max} T) + (1 - p_{\max}) C_g T.$
The first term captures the regret on rounds where $\Pi'$ follows $\Pi$ and we attribute regret to $\Pi$, while the second term bounds the regret from clipped actions.
\endproof

\section{Simulation experiment on mood dynamics in bipolar disorder}
\phantomsection
\label{bdSimulation}
This experiment uses data from a six-week self-monitoring study of 43 individuals with bipolar disorder \cite{bp2021}, who logged mood symptoms twice daily via the Lorevimo app. The digiBP survey (6 items) provided separate depressive and manic severity scores. The original study developed 84 intervention prompts based on Acceptance and Commitment Therapy (ACT), divided equally into three types—openness, awareness, and engagement—each targeting psychological flexibility through different mechanisms. In our simulation, these intervention types are modeled as discrete arms in a bandit framework, with rewards defined as reductions in depressive scores following intervention. DigiBP mood scores showed autoregressive patterns, justifying a linear state dynamics model with deterministic transitions and Gaussian noise for rewards \citep{bp2021}. We applied Expectation Maximization \citep{rabiner2002tutorial} to infer system parameters from longitudinal data, using the model described in Section~\ref{ToyReg} with three arms. Results averaged over 26 patients and 20 runs (each with 600 arm pulls, simulating ~10 months of monitoring) are :
\begin{figure}[H]
    \centering
    \begin{minipage}{0.35\textwidth}
        \centering
        \includegraphics[width=\textwidth]{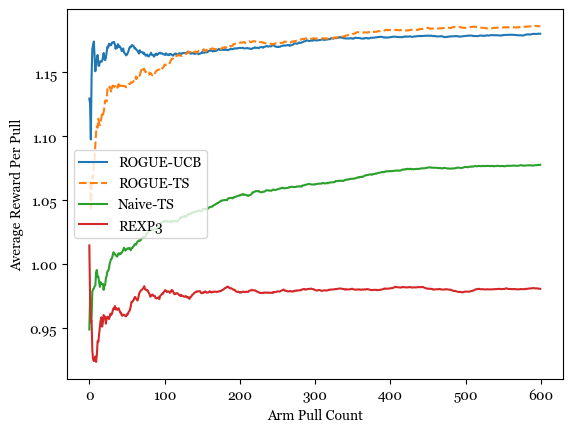}
        \caption{Average reward for Bipolar Disorder simulation}
        \label{fig:reward of BPS}
    \end{minipage}
    \begin{minipage}{0.35\textwidth}
        \centering
        \includegraphics[width=\textwidth]{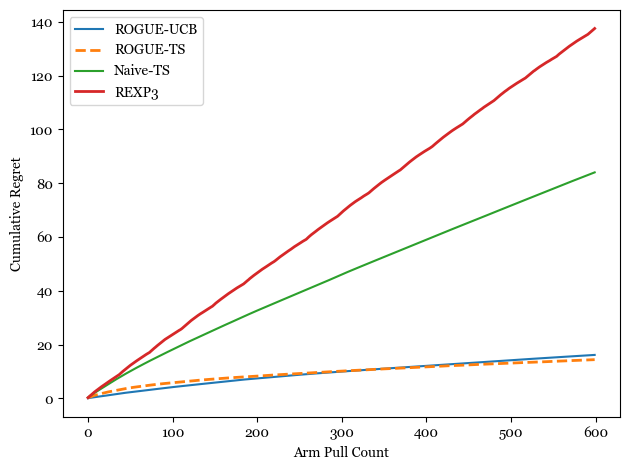}
        \caption{Cumulative regret for Bipolar Disorder simulation}
        \label{fig:regret of BPS}
    \end{minipage}
\end{figure}
The simulation results indicate that ROGUE-TS performs well in adapting to the evolving mood dynamics. Both ROGUE variants significantly outperform naive TS and REXP3, with naive TS performing marginally better than REXP3. Importantly, the improved reward achieved by the ROGUE algorithms translate to more effective intervention selections that are likely to result in greater reductions in depressive symptoms for patients. This suggests that adaptive algorithms incorporating personalized mood dynamics can enhance clinical decision-making by identifying the most helpful types of intervention prompts over time, ultimately supporting better patient outcomes.

\section{ROGUE-TS with Kalman Filtering}
\phantomsection
\label{app:kalman-rogue}
We present the full implementation of \textbf{ROGUE-TS with Kalman filtering}. Each arm maintains a posterior over the augmented state $ z_{a,t} = (\theta_a, x_{a,t}) $, and the model is reformulated to:
\begin{equation}
\begin{aligned}
z_{a,t+1} = 
\underbrace{\begin{bmatrix} I & 0 \\ 0 & A_a \end{bmatrix}}_{\widetilde{A}_a} z_{a,t} + 
\underbrace{\begin{bmatrix} 0 \\ \begin{bmatrix} B_a & K_a \end{bmatrix} \end{bmatrix}}_{\widetilde{B}_a} 
\begin{bmatrix} \pi_{a,t} \\ 1 \end{bmatrix}, \
r_{a,t} = \underbrace{\begin{bmatrix} \alpha_a^\top  \beta_a^\top \end{bmatrix}}_{H_a} z_{a,t} + \varepsilon_{a,t}, \quad \varepsilon_{a,t} \sim \mathcal{N}(0, \sigma_a^2)
\end{aligned}
\end{equation}

\begin{algorithm}[H]
\caption{ROGUE-TS with Kalman Filtering}
\label{alg:rogue-ts-kalman}
\begin{algorithmic}[1]
\Require Set of arms $\mathcal{A}$; initial priors $\mathcal{N}(\mu_{a,1}, \Sigma_{a,1})$ over $z_{a,1}$ for all $a \in \mathcal{A}$
\Statex \hspace{\algorithmicindent} Maintain posterior $(\mu_{a,t}, \Sigma_{a,t})$ for each $a$ at every $t$
\For{$t = 1$ to $T$}
    \For{each $a \in \mathcal{A}$}
        \State Sample $z_{a,t} \sim \mathcal{N}(\mu_{a,t}, \Sigma_{a,t})$, compute $\tilde{r}_{a,t} = H_a z_{a,t}$
    \EndFor
    \State Select $a_t = \arg\max_a \tilde{r}_{a,t}$, observe $r_{a_t,t}$

    \Statex \hspace{\algorithmicindent} \textbf{Correction (only for $a_t$):}
    \State $y_t = r_{a_t,t} - H_{a_t} \mu_{a_t,t}$,\quad $S_t = H_{a_t} \Sigma_{a_t,t} H_{a_t}^\top + \sigma_{a_t}^2,\quad K_t = \Sigma_{a_t,t} H_{a_t}^\top S_t^{-1}$
    \State $\mu_{a_t,t} = \mu_{a_t,t} + K_t y_t,\quad \Sigma_{a_t,t} = (I - K_t H_{a_t}) \Sigma_{a_t,t}$

    \Statex \hspace{\algorithmicindent} \textbf{Prediction step (all arms):}
    \For{each $a \in \mathcal{A}$}
        \State $\pi_{a,t}' \gets \begin{bmatrix} \pi_{a,t} & 1 \end{bmatrix}^\top$
        \State $\mu_{a,t+1} = \widetilde{A}_a \mu_{a,t} + \widetilde{B}_a \pi_{a,t}',\quad \Sigma_{a,t+1} = \widetilde{A}_a \Sigma_{a,t} \widetilde{A}_a^\top$
    \EndFor

\EndFor
\end{algorithmic}
\end{algorithm}

\end{APPENDICES}

\ACKNOWLEDGMENT{Acknowledgments}


\end{document}